\documentclass[journal]{new-aiaa} 
\usepackage[utf8]{inputenc}

\usepackage{graphicx}
\usepackage{amsmath}
\usepackage{bm}
\usepackage[version=4]{mhchem}
\usepackage{siunitx}
\usepackage{longtable,tabularx}
\usepackage[bf]{subfigure}

\allowdisplaybreaks

\title{Optimal Pose Estimation and Covariance Analysis with Simultaneous Localization and Mapping Applications}

\author{Saeed Maleki\footnote{Ph.D.~Graduate, Department of Mechanical and Aerospace Engineering, Email: saeedmal@buffalo.edu.}
}
\affil{University at Buffalo, State University of New York, Amherst, NY, 14260-4400}

\author{Adhiti Raman\footnote{Ph.D.~Candidate, Department of Automotive Engineering, Email: adhitir@g.clemson.edu.}
}
\affil{Clemson University, Clemson, SC, 29634}

\author{Yang Cheng\footnote{Associate Professor, Department of Aerospace Engineering, Email: cheng@ae.msstate.edu.  Associate Fellow AIAA.}}
\affil{Mississippi State University, Mississippi State, MS, 39762}

\author{John Crassidis\footnote{SUNY Distinguished Professor, Moog Endowed Chaired Professor of Innovation, Department of Mechanical and Aerospace Engineering, Email: johnc@buffalo.edu. Fellow AIAA.}}
\affil{University at Buffalo, State University of New York, Amherst, NY, 14260-4400}

\author{Matthias Schmid\footnote{Research Assistant Professor, Department of Automotive Engineering, Email: schmidm@clemson.edu.}}
\affil{Clemson University, Clemson, SC, 29634}

\begin{document}

\maketitle

\begin{abstract}
This work provides a theoretical analysis for optimally solving the pose estimation problem using total least squares for vector observations from landmark features, which is central to applications involving simultaneous localization and mapping. First, the optimization process is formulated  with observation vectors extracted from point-cloud features. Then, error-covariance expressions are derived. The attitude and position estimates obtained via the derived optimization process are proven to reach the bounds defined by the Cram\'er-Rao lower bound under the small-angle approximation of attitude errors. A fully populated observation noise-covariance matrix is assumed as the weight in the cost function to cover the most general case of the sensor uncertainty. This includes more generic correlations in the errors than previous cases involving an isotropic noise assumption. The proposed solution is verified using Monte Carlo simulations and an experiment with an actual LIDAR to validate the error-covariance analysis.

\end{abstract}

\section{Introduction}
Pose estimation is defined as estimating both the attitude and position (translation) of an object, which is central to simultaneous localization and mapping (SLAM) problems \cite{taketomi:17}. Sensors for pose estimation may either be passive, such as images taken by a camera, or active, such as radars or a light detection and ranging (LIDAR) sensor, in which the time delay between emission and return is measured. There are two basic categories of pose estimation: model-based and non-model based. Model-based approaches use {\it a priori} models of the object, and then estimate the pose of the object to match features from the model. Reference \cite{kelsey:06} states several various types of algorithms that can be used for model-based pose estimation, such as feature-based model tracking, model-to-image registration using various point-set registration algorithms, simultaneous pose and correspondence determination, template matching, contour tracking, and articulated object tracking. A survey of model-based pose estimation can be found in \cite{sala:14}. Non-model based approaches do not typically use any model attributes, but rather estimate an object's pose by separating the non-moving objects (e.g.~background scene) from the moving object to be tracked. Non-model pose estimation is typically accomplished by identifying, locating and tracking a number of features, such as a corner or an edge, on the object \cite{tae:19}. Both model and non-modeled based pose estimation can be used to simultaneously estimate either single object or multiple objects in a sensor's field-of-view (FOV), typical used for SLAM applications.

There are many applications of both model and non-model based pose estimation. These include spacecraft relative navigation \cite{christian:14,amico:14,kim:07} for space rendezvous and docking purposes \cite{singla:06}, absolute navigation of air vehicles from planar homographies \cite{merino:06}, relative air navigation \cite{fosbury:08} in GPS-denied environments \cite{izadi:15}, robotic navigation in unknown environments \cite{lu:97} and GPS-denied environments \cite{ma:12}, ground-vehicle position estimation using cameras \cite{barrois:09}, underwater-vehicle pose estimation using light beacons \cite{gracias:15}, and human pose estimation from images \cite{shotton:13}, including head position estimation \cite{chutorian:09}. The most basic steps for pose estimation first involve either using well-defined features extracted from an object or ``direct methods'' that not rely heavily on features. The former is usually used in conjunction with outlier elimination approaches, such as random sample consensus (RANSAC) \cite{fischler:81}, to provide more robustness to false matches. A number of algorithms exists for feature extraction, such as scale invariant feature transform (SIFT) \cite{lowe:04} or the speeded-up robust feature (SURF). One solution for the SLAM problem involves matching the histograms extracted from the features  \cite{rusu2008aligning}. Reference \cite{serafin2015nicp} utilizes a point as well as the local properties of its neighborhood, e.g.~the curvature and normal direction, to align point clouds.   Direct methods rely on ``template matching,'' and use a dense image alignment approach to refine the pose estimates that aligns a given target image to a sensed image (i.e.~image registration). The seminal algorithm by Lucas and Kanade, which outputs the required optical flow, is often used directly or as a basis for other approaches \cite{baker:04}. Once either feature extraction or direct methods have applied, pose estimation is then accomplished.

For registration of point clouds that are three-dimensional (3D) in nature, iterative closest point (ICP) algorithm implementations are commonly used \cite{rusinkiewicz:01}. However, it is well known that implementations of the ICP algorithm involve a large computational cost for arriving at the relative navigation estimates from the point clouds alone. Several researchers have implemented variants of the ICP algorithm that are more computationally efficient. Many of the variants are based on using a particular attitude parameterization \cite{shuster:93}. For example, \cite{crassidis:12} shows an approach that uses the Caley transform, which leads to a rigorously linear least-squares solution. But, because this algorithm is based on the classical Rodrigues parameters (Gibbs vector), a degree-of-freedom is lost for 180 degree rotations. An approach using modified Rodrigues parameters (MRP) is shown in \cite{bercovici:17}. Comparisons with Euler angles-based algorithms show that the MRP algorithm is more accurate and more computationally efficient. However, it is well known that MRPs are singular for 360 degree rotations. A quaternion formulation is shown in \cite{fathian:17}. The algorithm eliminates the unknown position vector by substraction of two images. Depths of the 3D point at each view are also eliminated by noting that they form a null vector for the matrix consisting of the point coordinates and their rotations. This leaves only the quaternion to be estimated, which is accomplished using the {\it q} method \cite{markley:14}.\footnote{Note that the authors of \cite{fathian:17} develop their own algorithm, which is equivalent to the {\it q} method.} Once the quaternion is determined, then the position vector and depths can be easily estimated. The advantage of the quaternion parameterization is that no singularities exist; however, a norm constraint must be satisfied. A dual-quaternion approach is shown in \cite{walker:91}. The advantages of using dual quaternions is that a single loss function can by minimized that is associated with the sum of the attitude and position errors, and no singular-condition rotations exist.

Most of the aforementioned registration algorithms do not employ a realistic sensor error-source model. For example, camera images are typically modeled using the classic pinhole model, which formulates the relationship between the coordinates of a point in 3D space and its projection onto the image plane of an ideal pinhole camera, leading to the ``collinearity equations.'' For mathematical convenience, the projected coordinates are put in a 3 $\times$ 1 unit-norm vector, which can assume that the focal length is 1 without loss in generality. Unfortunately, the measurement noise is also transformed when the collinearity equations are converted into the unit-vector form. A simple covariance model that is valid for small FOVs has been developed \cite{shuster:81}, called the QUEST measurement model (QMM), which is a singular matrix that arises for the unit-normalization of the observations. Its basic premise is that nearly all the probability of the errors is concentrated on a very small area about the direction of the unit vector, so the sphere containing that point can be approximated by a tangent plane. Reference \cite{cheng:06} expands upon the QMM for large-FOV cameras, which are typically used for many applications involving pose estimation. Both the QMM and large-FOV covariance models are singular, which may cause issues in estimation algorithms, but this can be overcome by replacing the singular matrix with a non-singular matrix using a rank-one update \cite{shuster:90,cheng:06}.

Properly accounting for the errors with the associated ICP equations leads to a total least squares (TLS) problem \cite{huffel:91}. TLS is similar to standard least squares, but errors are also added to the ``coefficient'' matrix, also known as the ``design'' matrix. Non-iterative solutions of TLS problems are only provided in special cases, such as when the weighting matrix is isotropic in nature. References \cite{ramos:97} and \cite{wen:05} show a TLS model for the ICP equations, but they assume that the errors have isotropic covariances. In \cite{estepar:04}, full weighting matrices are shown, and a generalized TLS-ICP is derived. However, in the aforementioned references, analytical expressions for the estimate error-covariance is not derived. The purpose of the present paper is to formulate the pose estimation problem for the most general sensor error case, including measurement correlations that may lead to a fully populated measurement matrix, which is the most general and realistic case. For isotropic errors, this paper also shows that the pose estimation problem is equivalent to a generalized version of Wahba's problem \cite{wahba:65}.  The associated estimate error-covariance matrices are also derived under the small error assumption. The error-covariance expressions are useful to quantify the overall performance of the TLS-ICP algorithms. Also, covariance expressions for the measurement residuals are derived in this paper. These expressions are useful to remove spurious measurements under realistic issues, such as mis-association of features in the pre-processing steps.

The outline of this paper is as follows. An introductory formulation of TLS is first shown. Next, a cost function is developed based on the pose estimation problem.
The necessary conditions for efficient pose estimation are then derived, and the cost function is written in an attitude-only format for the sake of simplicity in the proceeding derivations. Next, a linear approximation of the attitude error is derived to approximate the cost function within second-order terms in the attitude errors.
Efficient estimation for the attitude and translation vectors are then used to derive the error-covariance of the estimates and covariance of the measurement residuals, which are beneficial for control purposes in a sense that the most accurate pose estimates are provided to the control logic.
Finally, a numerical verification of the proposed TLS pose estimator is shown in the context of Monte Carlo simulations, as well as an experiment with actual LIDARs.

\begin{figure}[ht!]
  \centering
  \includegraphics[width=3in]{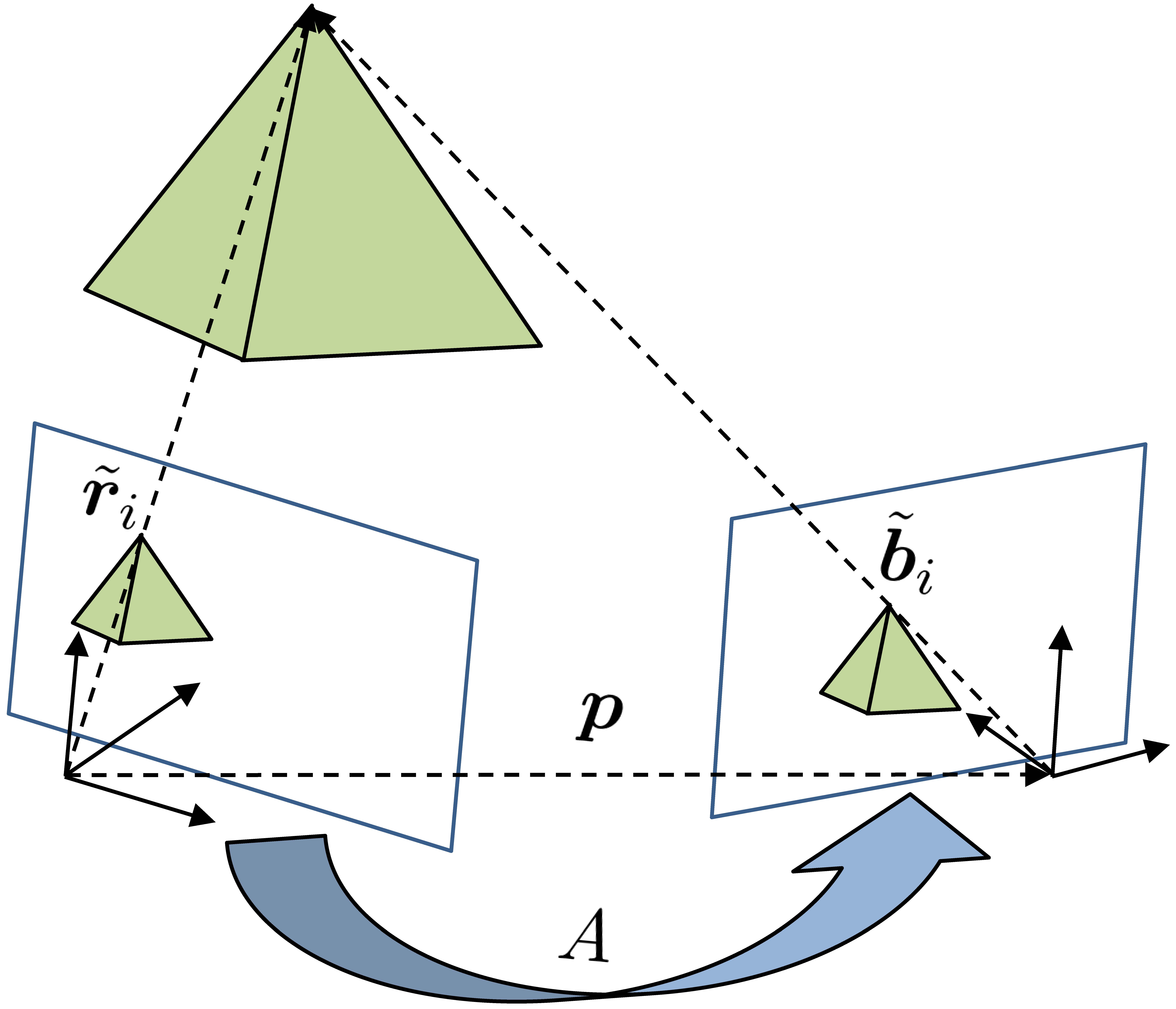}\\
  \caption{Geometric interpretation of the pose estimation problem.}\label{fig:geometry_calibration}
\end{figure}

\section{Problem Statement}
The pose estimation problem is described as finding the pose of a sensor attached to a vehicle with respect to another sensor or a reference frame.
The pose itself consists of two components: 1) a translation vector that connects the center of the two frames, and 2) an attitude matrix for the relative orientation of the vectors of the coordinate systems. A geometric model relating the pose of the reference frame to the body frame is shown in Fig.~\ref{fig:geometry_calibration}.
The associated geometric constraint acts as a measurement model. The vectors ${\tilde{\bm r}_i}$ and ${\tilde{\bm b}_i}$ are the coordinates of a point of interest, i.e.~an observation feature of a landmark in the environment, with respect to the reference and body frames, respectively, and $\bm{p}$ refers to the translation vector between the two coordinates.

The estimation approach provides an estimate of the attitude and the translation vector based on the measurement information.
The estimate error, which comprises the difference between the transformed version of ${\tilde{\bm r}_i}$ and ${\tilde{\bm b}_i}$, is given by
\begin{align}
    \bm{e}_i(\hat{A},\hat{\bm{p}})={\tilde{\bm b}_i}-\hat{A}{\tilde{\bm r}_i}+{\hat{\bm p}}
\end{align}
where $\hat{A}$ is the estimated attitude matrix, and $\hat{\bm{p}}$ is the estimated translation vector.
For the case of perfect measurements, the error samples $\bm{e}_i$ are all zero, and the problem can be solved with the measurement model of the form
\begin{equation}\label{perfect_measurement}
    \bm{b}_i=\hat{A}\bm{r}_i-\hat{\bm{p}}
\end{equation}
where $\bm{b}_i$ and $\bm{r}_i$ denote the true values of the observation vectors.
But in an actual applications, these errors are not zero, which leads to an optimization problem derived from a constrained maximum likelihood approach \cite{crassidis:12}, given by
\begin{equation}
\begin{gathered}
    \underset{\hat{A},{\hat{\bm p}}}{\min}\text{\ }J=\frac{1}{2}\sum_{i=1}^n\left({\tilde{\bm b}_i}-\hat{A}{\tilde{\bm r}_i}+\hat{\bm{p}}\right)^T R_i^{-1}\left({\tilde{\bm b}_i}-\hat{A}{\tilde{\bm r}_i}+{\hat{\bm p}}\right)\label{eqn:cost_att}\\
    \text{subject}\text{\ }\text{to:\ }\hat{A}^T \hat{A}= I_{3\times3} \text{\ },\text{\ }\det(\hat{A})=1
\end{gathered}
\end{equation}
where $I_{3\times3}$ is a $3 \times 3$ identity matrix, and $R_i$ is the measurement covariance matrix that accounts for errors in both $\tilde{\bm b}_i$ and $\tilde{\bm r}_i$, as well as correlations that exist between them.
The determinant condition is required so that $\hat{A}$ is a proper orthogonal matrix. 
The observation errors are defined as
\begin{subequations}
\begin{gather}
    \boldsymbol{\Delta} \bm{b}_i={\tilde{\bm b}}_i-\bm{b}_i\label{eqn:def_Delta_b}\\
    \boldsymbol{\Delta} \bm{r}_i={\tilde{\bm r}}_i-\bm{r}_i\label{eqn:def_Delta_r}
\end{gather}
\end{subequations}
It is assumed that zero-mean Gaussian measurement errors exist, with 
\begin{subequations}
\begin{gather}
    \mathcal{E}\{\boldsymbol{\Delta} \bm{b}_i\boldsymbol{\Delta} \bm{b}^T_i\}=R_{b_i}\label{eqn:cov_noise_bi}\\
    \mathcal{E}\{\boldsymbol{\Delta} \bm{r}_i\boldsymbol{\Delta} \bm{r}^T_i\}=R_{r_i}\label{eqn:cov_noise_ri}\\
    \mathcal{E}\{\boldsymbol{\Delta} \bm{r}_i\boldsymbol{\Delta} \bm{b}^T_i\}=\mathcal{E}\{\boldsymbol{\Delta} \bm{b}_i\boldsymbol{\Delta} \bm{r}^T_i\}^T=R_{rb_i}\label{eqn:cov_noise_rbi}\\
    R_i=\mathcal{E}\left\{\begin{bmatrix}\boldsymbol{\Delta} \bm{r}_i\\\boldsymbol{\Delta} \bm{b}_i\end{bmatrix}\begin{bmatrix}\boldsymbol{\Delta} \bm{r}^T_i&\boldsymbol{\Delta} \bm{b}^T_i\end{bmatrix}\right\}=\begin{bmatrix}R_{r_i}&R_{rb_i}\\R^T_{rb_i}& R_{b_i}\end{bmatrix}\label{eqn:R_i}
\end{gather}
\end{subequations}
where $\mathcal{E}$ denotes expectation. The optimization problem in Eq.~\eqref{eqn:cost_att} can be shown to be related to a total least squares (TLS) problem  \cite{cheng2021optimal}. 

This paper solves the pose estimation problem in Eq.~\eqref{eqn:cost_att} with the fully populated noise-covariance matrices $R_{r_i}$, $R_{b_i}$ and $R_{{rb}_i}$ using TLS.
Note that although there are nine components in the attitude matrix $A$, only three of them are independent, so the attitude estimation solution can be accomplished with a minimum of three parameters, which can be Euler angles or any other minimal attitude parameterization \cite{shuster:93}. 
Regardless of the attitude parameterization used in the optimization process, the estimated attitude can be related to the true attitude using an attitude error-vector involving small roll, pitch and yaw angles, as will be seen in Section \ref{tls_derivation}.

\subsection{Overview of Linear Least Squares and Total Least Squares}\label{intro_tls}
This section briefly introduces linear and total least squares, how they are related, and their differences.
For a more in-depth review of the TLS, the reader is referred to \cite{golub1973some, markovsky2007overview, golub1980analysis}. 

Consider the measurement model of the form 
\begin{equation}\label{eqn:meas_model_LS}
    \tilde{\bm{y}}=H\bm{x}+\boldsymbol{\Delta}\bm{y}
\end{equation}
where $H $ is an $m\times n$ deterministic design matrix with no errors, $\bm{x}$ is the $n\times 1$ vector of unknowns, $\tilde{\bm{y}}$ is the $m \times 1$ measurement vector, and $\boldsymbol{\Delta}\bm{y}$ is the $m \times 1$ measurement error-vector, 
The least squares estimate of $\bm{x}$ is given by solving the following problem:
\begin{equation}\label{cost_ls}
\begin{gathered}
    \underset{\hat{\bm{x}}}{\min}\text{\ }J=\frac{1}{2}\boldsymbol{\Delta}\bm{y}^T\boldsymbol{\Delta}\bm{y} \\
    \text{subject to: }\bm{\hat{y}}=H\hat{\bm{x}}
\end{gathered}
\end{equation}
The main underlying assumption in the statistical analysis of least squares is that $\tilde{\bm{y}}$ has a Gaussian distribution with the conditional likelihood function given by
\begin{equation}\label{eqn:liklihood_ls}
    p(\bm{\tilde{y}}|\bm{x})=\frac{1}{(2\pi)^{\frac{m}{2}}\big[\det(R_{yy})\big]^{\frac{1}{2}}}\text{exp}\left\{-\frac{1}{2}(\bm{\tilde{y}}-H \bm{x})^TR^{-1}_{yy}(\bm{\tilde{y}}-H \bm{x})\right\}
\end{equation}
where the distribution mean is denoted by $H\bm{x}$ and the covariance is $R_{yy}$. 
Because of the properties of the exponential function, maximizing the likelihood function in Eq.~\eqref{eqn:liklihood_ls} is equivalent to minimizing the negative of the log-likelihood. 
The estimate and its associated error-covariance are given by
\begin{subequations}
\begin{gather}
    \hat{\bm{x}}=\left(H^T R^{-1}_{yy} H\right)^{-1}H^TR^{-1}_{yy}\tilde{\bm{y}}\\
    \text{cov}\{\hat{\bm{x}}\}=\left(H^T R^{-1}_{yy} H\right)^{-1}
\end{gather}
\end{subequations}
One important property of linear least squares is that the estimate is unbiased \cite{crassidis:12}. 

As stated previously, the design matrix $H$ in the least squares measurement model in Eq.~\eqref{eqn:meas_model_LS} has no errors.
If this underlying assumption does not exist anymore, which happens in many applications, as will be seen in the SLAM problem in Section \ref{tls_derivation}, then another formulation must be used to consider the errors in the design matrix. This leads to the TLS problem, with parameters defined by

\begin{subequations}
\begin{gather}
  \tilde{\bm y}=\bm{y}+\boldsymbol{\Delta}\bm{y}\\
  \bm{y}=H\bm{x}\\
  \tilde{H}=H+\Delta{H}
\end{gather}
\end{subequations}
where  $\Delta H$ denotes the errors in the design matrix. Consider the following augmented matrix:
\begin{subequations}
\begin{gather}
    D=[H \,\,\, \bm{y}]
    \\
    \tilde{D}=[\tilde{H} \,\,\, \tilde{\bm{y}}]
\end{gather}
\end{subequations}
The conditional likelihood function of the TLS problem is defined by 
\begin{equation}
     p(\tilde{D}|D)=\frac{1}{(2\pi)^{\frac{m}{2}}\big[\det(R)\big]^{\frac{1}{2}}}\text{exp}\left\{-\frac{1}{2}\text{vec}(\tilde{D}-D)^TR^{-1}\text{vec}(\tilde{D}-D)\right\}
\end{equation}
where the vec operator stacks all columns of a matrix in a single column.
The maximum likelihood approach for this cost function leads to the minimization of the log-likelihood function as
\begin{equation}\label{eqn:cost_TLS_D}
\begin{gathered}
    J(\hat{D})=\frac{1}{2}\text{vec}(\tilde{D}-\hat{D})^T R^{-1} \text{vec}(\tilde{D}-\hat{D})\\
    \text{subject to}: \hat{D}\,{\hat{\bm z}}=\textbf{0} \\
\end{gathered}
\end{equation}
where $\hat{D}$ is the estimate of $D$, and $\hat{\bm{z}}= [{\hat{\bm x}}^T \,\,\, -1]^T$. A unique solution for this problem can be obtained if $\text{rank}(D)=n$.
Also, $R$ is the covariance matrix that accounts for the errors in both $\tilde{\bm y}$ and $\tilde{H}$. 
Although the TLS solution is known to be biased, the TLS problem is proven to reach the Cram\'er-Rao lower bound (CRLB) \cite{crassidis:12} for the estimate error-covariance to within first-order error-terms, and therefore is an {\it efficient} estimator \cite{crassidis2019maximum}.
Closed-form solutions for the TLS problem are possible only when $R$ is an isotropic matrix.

\subsection{Pose Estimation Sensor Model}\label{p_est_sensor_model}
The relation between the true vectors $\bm{b}_i$ and $\bm{r}_i$ is given by
\begin{equation}
\begin{split}
    \bm{b}_i&=A\bm{r}_i-\bm{p} \\
    &=\begin{bmatrix}\bm{r}^T_i & \bm{0}_{1\times3} & \bm{0}_{1\times3}\\
    \bm{0}_{1\times3} & \bm{r}^T_i & \bm{0}_{1\times3} \\
    \bm{0}_{1\times3} & \bm{0}_{1\times3} & \bm{r}^T_i  \end{bmatrix}\begin{bmatrix}\bm{a}_1\\
    \bm{a}_2\\
    \bm{a}_3\end{bmatrix}-\bm{p}\\
    &=H_i\bm{x}-\bm{p} \\ \label{eqn:det_tls_sensor_model}
    &\equiv\bm{y}_i
\end{split}
\end{equation}
The matrix $H_i$ is the individual sensor-model design matrix and $\bm{a}_i\ ,\  i=1,\,2,\,3$, are the columns of the attitude matrix $A$.
However, a perfect measurement model is not realistic because of noise in the design matrix, as well as the observation vectors of Eq.~\eqref{eqn:det_tls_sensor_model}.
Therefore, in the actual version of the sensor model in Eq.~\eqref{eqn:det_tls_sensor_model} the following relation is used:
\begin{equation}
\begin{split}
    {\tilde{\bm b}_i}-\boldsymbol{\Delta}\bm{b}_i&=A({\tilde{\bm r}_i}-\boldsymbol{\Delta}\bm{r}_i)-\bm{p}\\
    &=(\tilde{H}_i-\Delta H_i){\bm x}-\bm p \\
    &\equiv{\tilde{\bm y}_i}-\boldsymbol{\Delta} \bm{y}_i
\end{split}
\end{equation}
where the design matrix $\tilde{H}_i$ and the observation vector $\tilde{\bm{y}}_i$ have the errors of $\Delta H_i$ and $\boldsymbol{\Delta} \bm{y}_i$, respectively.
Because the model is linear in terms of the unknowns $\bm{x}$ and the translation vector $\bm{p}$, then the problem can be posed using a TLS formulation with the constraint 
\begin{equation}
 {\hat{\bm b}_i}=\hat{A}{\hat{\bm r}_i}-{\hat{\bm p}}
\end{equation}
which is equivalent to
\begin{equation}\label{eqn:dz=0}
\hat{D}_i{\hat{\bm z}}-{\hat{\bm p}}=\bm{0}    
\end{equation}
where $\hat{\bm{z}}= [{\hat{\bm x}}^T \,\,\, -1]^T$ and $\hat{D}_i=[\hat{H}_i  \,\,\, \bm{\hat{y}}_i]$.

\subsection{Total Least Squares Derivation for Pose Determination}\label{tls_derivation}
The TLS cost function is given by
\begin{equation}\label{eqn:original_J_tls}
    J(\hat{D}_i)=\frac{1}{2}\sum_{i=1}^n\text{vec}(\tilde{D}_i-\hat{D}_i)^T R_{D_i}^{-1} \text{vec}(\tilde{D}_i-\hat{D}_i)
\end{equation}
where $R_{D_i}$ is the covariance of $\text{vec}(\tilde{D}_i)$, and $n$ is the number of features in each sensor scan.
The augmented cost function that includes the linear constraint in Eq.~\eqref{eqn:dz=0} is given by
\begin{equation}
\begin{split}
    J_a(\hat{D}_i,\boldsymbol{\lambda}_i)&=\frac{1}{2}\sum_{i=1}^n\text{vec}(\tilde{D}_i-\hat{D}_i)^T R_{D_i}^{-1} \text{vec}(\tilde{D}_i-\hat{D}_i)+\boldsymbol{\lambda}_i^T\left(\hat{D}_i\hat{\bm{z}}-\hat{\bm{p}}\right)
\end{split}
\end{equation}
where $\boldsymbol{\lambda}_i$ is the Lagrange multiplier for the $i$th constraint. Taking the partial derivative of the augmented cost function with respect to $\text{vec}(\hat{D}_i)$, utilizing Eq.~\eqref{eqn:kronecker_vec}, and the necessary condition for the minimization problem gives
\begin{equation}\label{eqn:vecdhat}
    \text{vec}(\hat{D}_i)=\text{vec}(\tilde{D}_i)-R_{D_i}(\hat{\bm{z}}^T\otimes I_{3\times 3})^T\boldsymbol{\lambda}_i
\end{equation}
where $\otimes$ is the Kronecker product \cite{steeb2011matrix}. Using Eq.~\eqref{eqn:kronecker_vec}, the constraint in Eq.~\eqref{eqn:dz=0} can be written as
\begin{equation}\label{eqn:kron_constraint}
    (\hat{\bm{z}}^T\otimes I_{3\times 3})\text{vec}(\hat{D}_i)-\hat{\bm{p}}=\bm{0}
\end{equation}
Substituting Eq.~\eqref{eqn:vecdhat} into the constraint gives 
\begin{equation}
    (\hat{\bm{z}}^T\otimes I_{3\times 3})\left[\text{vec}(\tilde{D}_i)-R_{D_i}(\hat{\bm{z}}^T\otimes I_{3\times 3})^T\boldsymbol{\lambda}_i\right]-\hat{\bm{p}}=\bm{0}
\end{equation}
Solving for $\boldsymbol{\lambda}_i$ leads to
\begin{equation}\label{eqn:solvelambda}
    \boldsymbol{\lambda}_i=Q^{-1}_{\hat{\lambda}_i}\left[ (\hat{\bm{z}}^T\otimes I_{3\times 3})\text{vec}(\tilde{D}_i)-\hat{\bm{p}}\right]
\end{equation}
where $Q_{\hat{\lambda}_i}=(\hat{\bm{z}}^T\otimes I_{3\times 3})R_{D_i}(\hat{\bm{z}}^T\otimes I_{3\times 3})^T$.
Note that in this paper, $Q_{\hat{\lambda}_i}$ is considered to be a positive-definite matrix, though it might be singular in some sensor models, such as the QMM \cite{shuster:81}. 
This problem can be solved using a similar approach to the eigenvalue decomposition in \cite{amiri2017weighted}.
Using the necessary condition for $\hat{D}_i$ from Eq.~\eqref{eqn:vecdhat}, and substituting the Lagrange multiplier $\boldsymbol{\lambda}_i$ from Eq. \eqref{eqn:solvelambda} gives
\begin{equation}
    \begin{split}
        \text{vec}(\tilde{D}_i)-\text{vec}(\hat{D}_i)&=R_{D_i}(\hat{\bm{z}}^T\otimes I_{3\times 3})^TQ^{-1}_{\hat{\lambda}_i}\left[ (\hat{\bm{z}}^T\otimes I_{3\times 3})\text{vec}(\tilde{D}_i)-\hat{\bm{p}}\right]\\
    &=R_{D_i}(\hat{\bm{z}}^T\otimes I_{3\times 3})^TQ^{-1}_{\hat{\lambda}_i}\left(\tilde{D}_i\hat{\bm{z}}-\hat{\bm{p}}\right) \label{eqn:step2vecdd} 
    \end{split}
\end{equation}
 
Substituting Eq.~\eqref{eqn:step2vecdd} into the original cost function in Eq.~\eqref{eqn:original_J_tls} yields 
\begin{equation}
    J(\hat{\bm{z}},\hat{\bm{p}})=\frac{1}{2}\sum_{i=1}^n\left(\hat{\bm{p}}-\tilde{D}_i\hat{\bm{z}}\right)^T Q^{-1}_{\hat{\lambda}_i}\left(\hat{\bm{p}}-\tilde{D}_i\hat{\bm{z}}\right)
\end{equation}
This results in a formulation, satisfying the constraint in Eq.~\eqref{eqn:dz=0} in the cost function in Eq.~\eqref{eqn:original_J_tls}. Then in terms of the observation vectors $\tilde{\bm{b}}_i$ and $\tilde{\bm{r}}_i$, and from $\hat{\bm{p}}-\tilde{D}_i\hat{\bm{z}}=\tilde{\bm{b}}_i-\hat{A}\tilde{\bm{r}}_i+\hat{\bm{p}}$, the following expression is given:
\begin{equation}\label{eqn:cost_ahat_phat}
    J(\hat{A},\hat{\bm{p}})=
    \frac{1}{2}\sum_{i=1}^n\left(\tilde{\bm{b}}_i-\hat{A}\tilde{\bm{r}}_i+\hat{\bm{p}}\right)^T Q^{-1}_{\hat{\lambda}_i}\left(\tilde{\bm{b}}_i-\hat{A}\tilde{\bm{r}}_i+\hat{\bm{p}}\right)
\end{equation}
For simplicity in the proceeding derivations, the cost function can be written only in terms of the attitude matrix, which is known as the attitude-only cost function. For this purpose,  $\hat{\bm{p}}$ needs to be eliminated from the cost function in Eq.~\eqref{eqn:cost_ahat_phat}. The necessary condition for the translation vector $\hat{\bm{p}}$ results in
\begin{equation}\label{eqn:def_phat}
    \hat{\bm{p}}=-\left(\sum_{i=1}^{n}Q^{-1}_{\hat{\lambda_i}}\right)^{-1} \left[\sum_{i=1}^{n}Q^{-1}_{\hat{\lambda_i}}\left(\tilde{\bm{b}}_i-\hat{A}\tilde{\bm{r}}_i\right)\right]
\end{equation}
Substituting the optimal value of $\hat{\bm{p}}$ into Eq.~\eqref{eqn:cost_ahat_phat}, the attitude-only cost function becomes
\begin{equation}\label{eqn:cost_ahat}
    J(\hat{A})=\frac{1}{2}\Bigg[\sum_{i=1}^{n}\big(\tilde{\bm{b}}_i-\hat{A}\tilde{\bm{r}}_i\big)^T Q^{-1}_{\hat{\lambda_i}}\big(\tilde{\bm{b}}_i-\hat{A}\tilde{\bm{r}}_i\big)\Bigg]-\frac{1}{2}\Bigg[\sum_{i=1}^{n}Q^{-1}_{\hat{\lambda_i}}\big(\tilde{\bm{b}}_i-\hat{A}\tilde{\bm{r}}_i\big)\Bigg]^T\Bar{Q}_{\hat{\lambda}}\Bigg[\sum_{i=1}^{n}Q^{-1}_{\hat{\lambda_i}}\big(\tilde{\bm{b}}_i-\hat{A}\tilde{\bm{r}}_i\big)\Bigg]
\end{equation}
where
\begin{subequations}
    \begin{gather}
        \bm{\Tilde{b}}_i=\bm{b}_i+\boldsymbol{\Delta} \bm{b}_i\label{eqn:b_tilde_def}\\
    \bm{\Tilde{r}}_i=\bm{r}_i+\boldsymbol{\Delta} \bm{r}_i\label{eqn:r_tilde_def}
    \end{gather}
\end{subequations}
are the observation vectors with their corresponding covariance shown in Eqs.~\eqref{eqn:cov_noise_bi}, \eqref{eqn:cov_noise_ri} and \eqref{eqn:cov_noise_rbi}, and 
\begin{equation}
    \Bar{Q}_{\hat{\lambda}}=\left(\sum_{i=1}^{n}Q^{-1}_{\hat{\lambda_i}}\right)^{-1}\label{q_cap_bar}
\end{equation}
Also it is proven in \cite{cheng2021optimal} that the weight matrix $Q_{\hat{\lambda}_i}$ can be derived as a function of the attitude matrix and the observation noise covariance matrices as
\begin{equation}\label{eqn:q_lam_hat_i}
    Q_{\hat{\lambda}_i}=\hat{A}R_{r_i}\hat{A}^T-\hat{A}R_{rb_i}-R_{rb_i}^T\hat{A}^T+R_{b_i}
\end{equation}
 
\subsection{Covariance Analysis of the Estimates and Residuals}\label{cov_analysis}
The covariance expressions for the attitude as well as the translation and observation vector estimates are now derived.
Note that the cost function can be written in terms of the attitude error, defined by $\boldsymbol{\delta} \boldsymbol{\alpha}$, since only three independent components exist inside of the attitude matrix. The relation between the true and the estimated attitude matrix can be expressed as
\begin{equation}
    \hat{A}=\text{exp}(-[\boldsymbol{\delta} \boldsymbol{\alpha}\times])A
\end{equation}
where $[\cdot\times]$ denotes the cross product matrix of a vector, as shown by Eq.~\eqref{cross_prod_def}.
Using a small-angle assumption and a first-order approximation of the attitude error, the attitude estimate can be written as
\begin{equation}
    \hat{A}\approx \big(I_{3\times3}-[\boldsymbol{\delta} \boldsymbol{\alpha}\times ]\big)A\label{eqn:att_error_1st}
\end{equation}
The cost function needs be derived up to second-order in terms of the attitude error $\boldsymbol{\delta} \boldsymbol{\alpha}$, since  error-covariance expressions for a first-order approximation of the unknown errors are sought. The derivation begins with the approximation of the error-terms inside the cost function of Eq.~\eqref{eqn:cost_ahat}. The attitude approximation in Eq.~\eqref{eqn:att_error_1st}, and Eqs.~\eqref{eqn:cross_prod_def} and \eqref{eqn:kronecker_vec}, as well as Eqs.~\eqref{eqn:b_tilde_def} and \eqref{eqn:r_tilde_def}, are utilized for the formulation of the observation vectors, yielding 
\begin{equation}
\begin{split}
    \tilde{\bm{b}}_i-\hat{A}\tilde{\bm{r}}_i&=\bm{b}_i+\boldsymbol{\Delta}\bm{b}_i-\big(I_{3\times3}-[\boldsymbol{\delta} \boldsymbol{\alpha}\times]\big)A(\bm{r}_i+\boldsymbol{\Delta}\bm{r}_i)\\
    &\approx \bm{b}_i-A\bm{r}_i+\boldsymbol{\Delta} \bm{b}_i-A\boldsymbol{\Delta} \bm{r}_i - [A\bm{r}_i\times]\boldsymbol{\delta} \boldsymbol{\alpha}
\end{split}
\end{equation}
The following abbreviations are introduced for simplicity:
\begin{subequations}
    \begin{align}
        \boldsymbol{\Delta} \bm{a}_i&\equiv\boldsymbol{\Delta} \bm{b}_i-A\boldsymbol{\Delta} \bm{r}_i\label{eqn:def_Dai}\\
    \bm{p}&=A\bm{r}_i-\bm{b}_i\\
    \boldsymbol \nu_i &\equiv-\boldsymbol{\Delta} \bm{a}_i+\bm{p}\label{eqn:def_nu_i}\\
    \mathcal{A}_i&\equiv[A\bm{r}_i\times]
    \end{align}
\end{subequations}
This allows for the reformulation of the error-term $\tilde{\bm{b}}_i-\hat{A}\tilde{\bm{r}}_i$ as
\begin{align}
    \label{bti-ati}
    \tilde{\bm{b}}_i-\hat{A}\tilde{\bm{r}}_i\approx -\boldsymbol \nu_i-\mathcal{A}_i\boldsymbol{\delta} \boldsymbol{\alpha}
\end{align}
Note that $Q_{\hat{\lambda}_i}$ in Eq.~\eqref{eqn:q_lam_hat_i} is also a function of the attitude estimate $\hat{A}$ and subsequently of the attitude error $\boldsymbol{\delta} \boldsymbol{\alpha}$. As the neighboring terms are already a function of the first-order attitude error, any other term besides the attitude-error dependent ones in $Q_{\hat{\lambda}_i}$ (terms that are independent of $\boldsymbol{\delta} \boldsymbol{\alpha}$) are not kept for the second-order approximation of the cost function. The matrix $Q_{\hat{\lambda}_i}$ is now written as
\begin{equation}
    \begin{split}
        \label{eqn:expand_Q}
    Q_{\hat{\lambda}_i}&=\hat{A}R_{r_i}\hat{A}^T-\hat{A}R_{rb_i}-R_{rb_i}^T\hat{A}^T+R_{b_i} \\
    &=\big(I_{3\times3}-[\boldsymbol{\delta} \boldsymbol{\alpha}\times]\big)A R_{r_i} A^T \big(I_{3\times3}-[\boldsymbol{\delta} \boldsymbol{\alpha}\times]\big)^T -\big(I_{3\times3}-[\boldsymbol{\delta} \boldsymbol{\alpha}\times]\big)A R_{rb_i}-R_{rb_i}^T A^T (I_{3\times3}-[\boldsymbol{\delta} \boldsymbol{\alpha}\times])^T +R_{b_i}
    \end{split}
\end{equation}
Further decomposition of $Q_{\hat{\lambda}_i}$ in terms of $\boldsymbol{\delta} \boldsymbol{\alpha}$ then yields
\begin{equation}
    Q_{\hat{\lambda}_i}=Q_{\lambda_i}+\delta Q_{\lambda_i}+\delta^2Q_{\lambda_i}\label{eqn:q_lam_i}
\end{equation}
where
\begin{subequations}
    \begin{gather}
    Q_{\lambda_i}=AR_{r_i}A^T-AR_{rb_i}-R_{rb_i}^TA^T+R_{b_i}\\
    \delta Q_{\lambda_i}=\mathcal{K}_i[\boldsymbol{\delta}\boldsymbol{\alpha}\times]+[\boldsymbol{\delta}\boldsymbol{\alpha}\times]^T\mathcal{K}_i^T\\
    \delta^2Q_{\lambda_i}=[\boldsymbol{\delta}\boldsymbol{\alpha}\times]AR_{r_i}A^T[\boldsymbol{\delta}\boldsymbol{\alpha}\times]^T
    \end{gather}
\end{subequations}
with
\begin{equation}
    \mathcal{K}_i=AR_{r_i}A^T-R_{rb_i}^TA^T
\end{equation}
The inverse of $Q_{\hat{\lambda}_i}$ is approximated by
\begin{align}
    \label{iQ_hat}
    Q^{-1}_{\hat{\lambda}_i}\approx Q^{-1}_{\lambda_i}-Q^{-1}_{\lambda_i}\delta Q_{\lambda_i}Q^{-1}_{\lambda_i}-Q^{-1}_{\lambda_i}\delta^2Q_{\lambda_i}Q^{-1}_{\lambda_i}
\end{align}
Therefore, the matrix $Q^{-1}_{\lambda_i}$ is the only portion of $Q^{-1}_{\hat{\lambda}_i}$ that is not a function of the attitude error.
This portion will be used for building the second-order cost function.
The second summation of the cost function in Eq.~\eqref{eqn:cost_ahat} contains $\Bar{Q}_{\hat{\lambda}}$, which itself is a function of $\boldsymbol{\delta \alpha}$.
The portion of $\Bar{Q}_{\hat{\lambda}}$ that does not depend on the attitude error needs to be extracted.
This way, the corresponding part of the cost function ignores higher-order terms.
The expansion of $\Bar{Q}_{\hat{\lambda}}$ is given as
\begin{align}
    \Bar{Q}_{\hat{\lambda}}=\Bigg(\sum_{i=1}^{n}Q^{-1}_{\lambda_i}\Bigg)^{-1}+\Bigg(\sum_{i=1}^{n}Q^{-1}_{\lambda_i}\Bigg)^{-1}\Bigg[\sum_{i=1}^n Q^{-1}_{\lambda_i}\bigg(\delta Q^{-1}_{\lambda_i}+\delta^2 Q^{-1}_{\lambda_i}\bigg)Q^{-1}_{\lambda_i}\Bigg]\Bigg(\sum_{i=1}^{n}Q^{-1}_{\lambda_i}\Bigg)^{-1}
\end{align}
The summations in the above equation can be abbreviated as
\begin{subequations}
    \begin{gather}
        S_{\lambda}=\Bigg(\sum_{i=1}^{n}Q^{-1}_{\lambda_i}\Bigg)^{-1}\label{eqn:s_lambda}
        \\
    \delta S_{\lambda}=S_{\lambda}\Bigg(\sum_{i=1}^n Q^{-1}_{\lambda_i}\delta Q^{-1}_{\lambda_i}Q^{-1}_{\lambda_i}\Bigg) S_{\lambda}
    \\
    \delta^2 S_{\lambda}=S_{\lambda}\Bigg(\sum_{i=1}^n Q^{-1}_{\lambda_i}\delta^2 Q^{-1}_{\lambda_i}Q^{-1}_{\lambda_i}\Bigg) S_{\lambda}
    \end{gather}
\end{subequations}
Hence, $S_{\lambda}$ emerges as the only term that is not a function of the attitude error. 
Utilizing the first-order errors in Eq.~\eqref{bti-ati}, the matrix $Q_{\lambda_i}$ in Eq.~\eqref{eqn:q_lam_i}, and $S_{\lambda}$ in Eq.~\eqref{eqn:s_lambda}, the following approximation up to second order of the cost function is given:
\begin{equation}
J(\boldsymbol{\delta}\boldsymbol{\alpha},\boldsymbol{\delta}\boldsymbol{\alpha}^T\boldsymbol{\delta}\boldsymbol{\alpha})\approx\frac{1}{2}\left(\boldsymbol{\delta} \boldsymbol{\alpha}^T\bm{g} + \boldsymbol{\delta} \boldsymbol{\alpha}^T \mathcal{H} \boldsymbol{\delta} \boldsymbol{\alpha}\right)\label{eqn:def_sec_order_cost}
\end{equation}
with
\begin{subequations}
    \begin{gather}
        \bm{g}= \left(\sum_{i=1}^n \mathcal{A}_i Q^{-1}_{\lambda_i} \right)S_{\lambda}\left(\sum_{i=1}^n Q^{-1}_{\lambda_i}\boldsymbol{\nu}_i \right)+\sum_{i=1}^n \mathcal{A}^T_i Q^{-1}_{\lambda_i}\boldsymbol{\nu}_i\label{def_G}\\
    \mathcal{H}=\left( \sum_{i=1}^n \mathcal{A}^T_i Q^{-1}_{\lambda_i}\mathcal{A}_i\right)-\left(\sum_{i=1}^n Q^{-1}_{\lambda_i}\mathcal{A}_i \right)^T  S_{\lambda} \left(\sum_{i=1}^n Q^{-1}_{\lambda_i}\mathcal{A}_i\right)\label{H_def}
    \end{gather}
\end{subequations}

From the necessary condition for the extremum of the cost function $J$ with respect to $\boldsymbol{\delta} \boldsymbol{\alpha}$, i.e.
\begin{align}
    \frac{\partial J}{\partial \boldsymbol{\delta} \boldsymbol{\alpha}}
    =\bm{0}
\end{align}
the vector estimate of the attitude-error emanates as
\begin{align}
    \boldsymbol{\delta} \boldsymbol{\alpha}=-\mathcal{H}^{-1}\bm{g}
\end{align}
Employing Eq.~\eqref{eqn:def_nu_i} allows for expanding the above expression to
\begin{equation}
    \boldsymbol{\delta} \boldsymbol{\alpha}=-\mathcal{H}^{-1}\left[\left(\sum_{i=1}^n \mathcal{A}_i Q^{-1}_{\lambda_i} \right)S_{\lambda}\bigg[\sum_{i=1}^n Q^{-1}_{\lambda_i}(\bm{p}-\boldsymbol{\Delta} \bm{a}_i) \bigg]+\sum_{i=1}^n \mathcal{A}^{T}_i Q^{-1}_{\lambda_i}(\bm{p}-\boldsymbol{\Delta} \bm{a}_i)\right]
\end{equation}
At the same time, the following relation holds:
\begin{equation}
    \left(\sum_{i=1}^n \mathcal{A}_i Q^{-1}_{\lambda_i} \right)S_{\lambda}\left(\sum_{i=1}^n Q^{-1}_{\lambda_i}\bm{p} \right)+\sum_{i=1}^n \mathcal{A}^{T}_i Q^{-1}_{\lambda_i}\bm{p}=\bm{0}
\end{equation}
Given that $\mathcal{A}^T_i=-\mathcal{A}_i$, as well as that $Q^{-1}_{\lambda_i}$, $S_{\lambda}$ and $\mathcal{H}$ are symmetric, the terms containing the translation vector $\bm{p}$ cancel, thus yielding
\begin{align}\label{eqn:d_alpha}
    \boldsymbol{\delta} \boldsymbol{\alpha}=\mathcal{H}^{-1}\left[\left(\sum_{i=1}^n \mathcal{A}_i Q^{-1}_{\lambda_i} \right)S_{\lambda}\left(\sum_{i=1}^n Q^{-1}_{\lambda_i}\boldsymbol{\Delta} \bm{a}_i \right)-\sum_{i=1}^n \mathcal{A}_i Q^{-1}_{\lambda_i}\boldsymbol{\Delta} \bm{a}_i\right]
\end{align}
This attitude error now allows for the derivation of the error-covariance expressions.
The estimation error-covariance of the attitude is defined as
\begin{equation}
\begin{split}
    P_{\boldsymbol{\delta} \boldsymbol{\alpha}}&=\mathcal{E}\{\boldsymbol{\delta} \boldsymbol{\alpha}\boldsymbol{\delta} \boldsymbol{\alpha}^T\}\\
    &=\mathcal{E}\Bigg\{\mathcal{H}^{-1}\left[\big(\sum_{i=1}^n \mathcal{A}_i Q^{-1}_{\lambda_i} \big)S_{\lambda}\big(\sum_{i=1}^n Q^{-1}_{\lambda_i}\boldsymbol{\Delta} \bm{a}_i \big)-\sum_{i=1}^n \mathcal{A}_i Q^{-1}_{\lambda_i}\boldsymbol{\Delta} \bm{a}_i\right]
    \\
    & \times\left[\big(\sum_{i=1}^n \mathcal{A}_i Q^{-1}_{\lambda_i} \big)S_{\lambda}\big(\sum_{i=1}^n Q^{-1}_{\lambda_i}\boldsymbol{\Delta} \bm{a}_i \big)-\sum_{i=1}^n \mathcal{A}_i Q^{-1}_{\lambda_i}\boldsymbol{\Delta} \bm{a}_i\right]^T\mathcal{H}^{-T}\Bigg\}
\end{split}
\end{equation}
Further expansion of the individual terms leads to
\begin{equation}
\begin{split}
    P_{\boldsymbol{\delta} \boldsymbol{\alpha}}& =\mathcal{H}^{-1}\Bigg[-\sum_{i=1}^n\big(\mathcal{A}_i Q^{-1}_{\lambda_i}\mathcal{E}\{\boldsymbol{\Delta} \bm{a}_i\boldsymbol{\Delta} \bm{a}^T_i\}Q^{-1}_{\lambda_i}\mathcal{A}_i\big)\\
    &+\big(\sum_{i=1}^n\mathcal{A}_i Q^{-1}_{\lambda_i}\big)S_{\lambda}\big(\sum_{i=1}^nQ^{-1}_{\lambda_i}\mathcal{E}\{\boldsymbol{\Delta} \bm{a}_i\boldsymbol{\Delta} \bm{a}^T_i\}Q^{-1}_{\lambda_i}\big) S_{\lambda}\big(\sum_{i=1}^n Q^{-1}_{\lambda_i}\mathcal{A}^T_i\big)\\
    &-\big(\sum_{i=1}^n\mathcal{A}_iQ^{-1}_{\lambda_i}\mathcal{E}\{\boldsymbol{\Delta} \bm{a}_i\boldsymbol{\Delta} \bm{a}^T_i\}Q^{-1}_{\lambda_i}\big)S_{\lambda}\big(\sum_{i=1}^n Q^{-1}_{\lambda_i}\mathcal{A}^T_i\big)\\
    &+\big(\sum_{i=1}^n\mathcal{A}_i Q^{-1}_{\lambda_i}\big)S_{\lambda} \big(\sum_{i=1}^n Q^{-1}_{\lambda_i} \mathcal{E}\{\boldsymbol{\Delta} \bm{a}_i\boldsymbol{\Delta} \bm{a}^T_i\}Q^{-1}_{\lambda_i}\mathcal{A}_i\big)  \Bigg]\mathcal{H}^{-T}
\end{split}
\end{equation}
Employing the fact that
\begin{subequations}
    \begin{gather}
        P_{\boldsymbol{\Delta} \bm{a}_i}\equiv \mathcal{E}\{\boldsymbol{\Delta} \bm{a}_i\boldsymbol{\Delta} \bm{a}^T_i\}=Q_{\lambda_i}\label{eqn:P_delta_ai}\\
    \mathcal{E}\{\boldsymbol{\Delta} \bm{a}_i\boldsymbol{\Delta} \bm{a}^T_j\}=0_{3\times3}\ , j\neq i\label{eqn:P_delta_ainot}
    \end{gather}
\end{subequations}
the attitude error-covariance can be rewritten as
\begin{equation}
\begin{split}
    P_{\boldsymbol{\delta} \boldsymbol{\alpha}}&=\mathcal{H}^{-1}\Bigg[-\sum_{i=1}^n\big(\mathcal{A}_iQ^{-1}_{\lambda_i}Q_{\lambda_i}Q^{-1}_{\lambda_i}\mathcal{A}_i\big) \\
    &+\big(\sum_{i=1}^n\mathcal{A}_i Q^{-1}_{\lambda_i}\big)S_{\lambda}\big(\sum_{i=1}^n Q^{-1}_{\lambda_i} Q_{\lambda_i}Q^{-1}_{\lambda_i}\big)S_{\lambda}\big(\sum_{i=1}^n Q^{-1}_{\lambda_i}\mathcal{A}^T_i\big) \\
    &-\big(\sum_{i=1}^n\mathcal{A}_iQ^{-1}_{\lambda_i}Q_{\lambda_i}Q^{-1}_{\lambda_i}\big)S_{\lambda}\big(\sum_{i=1}^n Q^{-1}_{\lambda_i}\mathcal{A}^T_i\big)\\
    &+\big(\sum_{i=1}^n\mathcal{A}_i Q^{-1}_{\lambda_i}\big)S_{\lambda} \big(\sum_{i=1}^n Q^{-1}_{\lambda_i} Q_{\lambda_i}Q^{-1}_{\lambda_i}\mathcal{A}_i\big)  \Bigg]\mathcal{H}^{-T}
\end{split}
\end{equation}
The last two contributions in the above sum cancel out each other, thus yielding
\begin{equation}
    P_{\boldsymbol{\delta} \boldsymbol{\alpha}}=\mathcal{H}^{-1}\Bigg[-\sum_{i=1}^n\big(\mathcal{A}_iQ^{-1}_{\lambda_i}\mathcal{A}_i\big)+\big(\sum_{i=1}^n\mathcal{A}_i Q^{-1}_{\lambda_i}\big)S_{\lambda}\big(\sum_{i=1}^n Q^{-1}_{\lambda_i}\big)S_{\lambda}\big(\sum_{i=1}^n Q^{-1}_{\lambda_i}\mathcal{A}^T_i\big)  \Bigg]\mathcal{H}^{-T}
\end{equation}
With Eqs.~\eqref{eqn:s_lambda} and \eqref{H_def}, the above expression further simplifies to
\begin{equation}
    P_{\boldsymbol{\delta} \boldsymbol{\alpha}}=\mathcal{H}^{-1}\label{eqn:P_delta_alpha_final}
\end{equation}
which verifies that the estimate error-covariance of the attitude error is equal to the Hessian of the cost function. Note that a more detailed discussion of this observation is provided later in the context of the Fisher information matrix (FIM) for the cost function in Eq.~\eqref{eqn:cost_ahat}. 

The estimation error for the translation vector is now derived, denoted by  $\boldsymbol{\delta} \bm{p}$, begins by using 
\begin{equation}
    \hat{\bm{p}}=\bm{p}+\boldsymbol{\delta} \bm{p}\label{phat-p}
\end{equation}
Decomposition of Eq.~\eqref{eqn:def_phat} by utilizing Eq.~\eqref{bti-ati} to separate the first-order terms in the attitude error yields
\begin{equation}
    \bm{p}+\boldsymbol{\delta} \bm{p}\approx-\left(\sum_{i=1}^{n}Q^{-1}_{\lambda_i}\right)^{-1} \left[\sum_{i=1}^{n}Q^{-1}_{\lambda_i}\left(-\boldsymbol \nu_i-\mathcal{A}_i\boldsymbol{\delta} \boldsymbol{\alpha}\right)\right]
\end{equation}
Thus, the estimate-error $\boldsymbol{\delta} \bm{p}$ emerges as
\begin{equation}
    \boldsymbol{\delta} \bm{p}=-S_{\lambda}\left(\sum_{i=1}^{n}Q^{-1}_{\lambda_i}\left(-\boldsymbol \nu_i-\mathcal{A}_i\boldsymbol{\delta} \boldsymbol{\alpha}+\bm{p}\right)\right)
\end{equation}
Using the definition of $\boldsymbol \nu_i$ in Eq.~\eqref{eqn:def_nu_i} gives
\begin{equation}\label{eqn:err_p}
    \boldsymbol{\delta} \bm{p}=-S_{\lambda}\sum_{i=1}^{n}Q^{-1}_{\lambda_i}\left(\boldsymbol{\Delta} \bm{a}_i-\mathcal{A}_i\boldsymbol{\delta} \boldsymbol{\alpha}\right)
\end{equation}
The error-covariance of the translation vector within first-order of the estimation errors is given by
\begin{equation}
    \begin{split}
        \label{def_cov_phat}
        \text{cov}\{\hat{\bm{p}}\}&\equiv \mathcal{E}\left\{(\hat{\bm{p}}-\mathcal{E}\{\hat{\bm{p}}\})(\hat{\bm{p}}-\mathcal{E}\{\hat{\bm{p}}\})^T\right\}\\
        &=\mathcal{E}\{\boldsymbol{\delta} \bm{p}\boldsymbol{\delta} \bm{p}^T\}
    \end{split}
\end{equation}
where the fact that the estimate for the translation vector is unbiased within the first-order terms of error is used, and $\mathcal{E}\{\hat{\bm{p}}\}=\bm{p}$. 
Then, the translation vector error-covariance is computed as
\begin{equation}
    \text{cov}\{\hat{\bm{p}}\}=\mathcal{E}\left\{\left[-S_{\lambda}\sum_{i=1}^{n}Q^{-1}_{\lambda_i}\left(\boldsymbol{\Delta}\bm{a}_i-\mathcal{A}_i\boldsymbol{\delta} \boldsymbol{\alpha}\right)\right]\left[-S_{\lambda}\sum_{i=1}^{n}Q^{-1}_{\lambda_i}\left(\boldsymbol{\Delta}\bm{a}_i-\mathcal{A}_i\boldsymbol{\delta} \boldsymbol{\alpha}\right)\right]^T\right\}
\end{equation}
It is observed that the cross-covariance of the attitude errors and $\boldsymbol{\Delta} \bm{a}_i$ is required for the translation error-covariance, which is computed as
\begin{equation}
    \mathcal{E}\{\boldsymbol{\delta}\boldsymbol{\alpha}\Delta\bm{a}^T_i\}=\mathcal{H}^{-1}(\mathcal{A}_i-\Bar{\mathcal{A}})\label{eqn:p_alpha_ai}
\end{equation}
where 
\begin{align}
    \Bar{\mathcal{A}}=S_{\lambda}\sum_{i=1}^n Q^{-1}_{\lambda_i} \mathcal{A}_i
\end{align}
Using Eqs.~\eqref{eqn:P_delta_ai}, \eqref{eqn:P_delta_ainot} and \eqref{eqn:p_alpha_ai}, and from the attitude estimate error-covariance in Eq.~\eqref{eqn:P_delta_alpha_final}, the translation vector error-covariance is now given by
\begin{align}
    \text{cov}\{\hat{\bm{p}}\}=S_{\lambda}+\Bar{\mathcal{A}}P_{\boldsymbol{\delta}\alpha}\Bar{\mathcal{A}}^T\label{calc_cov_phat}
\end{align}

Because there are estimates for the observation vectors from the TLS formulation in Eq.~\eqref{eqn:vecdhat}, their associated covariance expressions can be derived. First an expression for their corresponding first-order residuals is found, and then this residual approximation is used to construct an analytical covariance formulation. Using Eq.~\eqref{eqn:vecdhat} for the observation vectors and using the derivation in \cite{cheng2021optimal}, it can be shown that the estimates of the observation vectors are
\begin{subequations}
    \begin{gather}
        \hat{\bm{b}}_i=\tilde{\bm{b}}_i+(R_{rb_i}^T\hat{A}^T-R_{b_i})Q^{-1}_{\hat{\lambda_i}}(\tilde{\bm{b}}_i-\hat{A}\tilde{\bm{r}}_i+\hat{\bm{p}}) \label{eqn:bhat-btilde}\\
    \hat{\bm{r}}_i=\tilde{\bm{r}}_i+(R_{r_i}\hat{A}^T-R_{rb_i})Q^{-1}_{\hat{\lambda_i}}(\tilde{\bm{b}}_i-\hat{A}\tilde{\bm{r}}_i+\hat{\bm{p}}) \label{eqn:rhat-rtilde}
    \end{gather}
\end{subequations}
Define the following estimate errors for the observation vectors:
\begin{subequations}
    \begin{gather}
        \boldsymbol{\delta}\bm{b}_i=\hat{\bm{b}}_i-\bm{b}_i\\
    \boldsymbol{\delta}\bm{r}_i=\hat{\bm{r}}_i-\bm{r}_i
    \end{gather}
\end{subequations}
The residual errors using Eq.~\eqref{eqn:r_tilde_def} and \eqref{eqn:b_tilde_def} are $\boldsymbol{\delta}\bm{b}_i-\boldsymbol{\Delta}\bm{b}_i$ and $\boldsymbol{\delta}\bm{r}_i-\boldsymbol{\Delta}\bm{r}_i$. Then deducting both sides of Eq.~\eqref{eqn:bhat-btilde} by $\bm{b}_i$ and Eq.~\eqref{eqn:rhat-rtilde} by $\bm{r}_i$ leads to
\begin{subequations}
    \begin{gather}
       \boldsymbol{\delta}\bm{b}_i-\boldsymbol{\Delta}\bm{b}_i=(R_{rb_i}^T\hat{A}^T-R_{b_i})Q^{-1}_{\hat{\lambda_i}}(\tilde{\bm{b}}_i-\hat{A}\tilde{\bm{r}}_i+\hat{\bm{p}})\\
    \boldsymbol{\delta}\bm{r}_i-\boldsymbol{\Delta}\bm{r}_i=(R_{r_i}\hat{A}^T-R_{rb_i})Q^{-1}_{\hat{\lambda_i}}(\tilde{\bm{b}}_i-\hat{A}\tilde{\bm{r}}_i+\hat{\bm{p}}) 
    \end{gather}
\end{subequations}
Their corresponding first-order approximations are given by
\begin{subequations}
    \begin{gather}
        \boldsymbol{\delta}\bm{b}_i-\boldsymbol{\Delta}\bm{b}_i\approx C_i(\boldsymbol{\Delta}\bm{a}_i-\mathcal{G}_i\boldsymbol{\delta}\bm{f})\label{apprxb-btilde}\\
    \boldsymbol{\delta}\bm{r}_i-\boldsymbol{\Delta}\bm{r}_i\approx D_i(\boldsymbol{\Delta}\bm{a}_i-\mathcal{G}_i\boldsymbol{\delta}\bm{f})\label{apprxr-rtilde}
    \end{gather}
\end{subequations}
where
\begin{subequations}
    \begin{align}
        C_i&=(R_{rb_i}^T A^T-R_{b_i})Q^{-1}_{\lambda_i}\\
    D_i&=(R_{r_i} A^T-R_{rb_i})Q^{-1}_{\lambda_i}\\
    \mathcal{G}_i&=\begin{bmatrix}\mathcal{A}_i &  -I_{3\times3}\end{bmatrix}\label{eqn:def_Gi}\\
    \boldsymbol{\delta}\bm{f}&=\begin{bmatrix}\boldsymbol{\delta}\boldsymbol{\alpha} \\ \boldsymbol{\delta} \bm{p}\end{bmatrix}_{6\times 1}\label{eqn:def_delta_f}
    \end{align}
\end{subequations}
Using
\begin{subequations}
    \begin{gather}
        P_{\boldsymbol{\Delta} \bm{a}_i}=Q_{\lambda_i}\\
        P_{\boldsymbol{\Delta}\bm{a}_i \boldsymbol{\Delta}\Bar{\bm{a}}}=S_{\lambda}\\
        P_{ \boldsymbol{\Delta}\Bar{\bm{a}} \boldsymbol{\delta}\boldsymbol{\alpha}} =0_{3\times 3}
    \end{gather}
\end{subequations}
 and from Eq.~\eqref{eqn:p_alpha_ai}, the covariances of the measurement residuals are given by
 \begin{subequations}
     \begin{align}
         \text{cov}(\hat{\bm{b}}_i-\tilde{\bm{b}}_i)&=\mathcal{E}\{(\boldsymbol{\delta}\bm{b}_i-\boldsymbol{\Delta}\bm{b}_i)(\boldsymbol{\delta}\bm{b}_i-\boldsymbol{\Delta}\bm{b}_i)^T\}\nonumber\\
    &=C_i (Q_{\lambda_i}-\mathcal{G}_i P_{\bm{f}} \mathcal{G}_i^T) C^T_i\label{eqn:cov_bi}\\
    \text{cov}(\hat{\bm{r}}_i-\tilde{\bm{r}}_i)&=\mathcal{E}\{(\boldsymbol{\delta}\bm{r}_i-\boldsymbol{\Delta}\bm{r}_i)(\boldsymbol{\delta}\bm{r}_i-\boldsymbol{\Delta}\bm{r}_i)^T\}\nonumber\\
    &=D_i (Q_{\lambda_i}-\mathcal{G}_i P_{\bm{f}} \mathcal{G}_i^T) D^T_i\label{eqn:cov_ri}
     \end{align}
 \end{subequations}
where
\begin{equation}
        P_{\bm{f}}\equiv\text{cov}\{\boldsymbol{\delta}\bm{f}\}=\left(\sum_{i=1}^n \mathcal{G}_i^T Q^{-1}_{\lambda_i} \mathcal{G}_i\right)^{-1}
\end{equation}
Equations\eqref{apprxb-btilde} and \eqref{apprxr-rtilde} can now be written as
\begin{subequations}
    \begin{gather}
        \boldsymbol{\delta}\bm{b}_i\approx\boldsymbol{\Delta}\bm{b}_i+ C_i(\boldsymbol{\Delta}\bm{a}_i-\mathcal{G}_i\boldsymbol{\delta}\bm{f})\\
    \boldsymbol{\delta}\bm{r}_i\approx\boldsymbol{\Delta}\bm{r}_i+ D_i(\boldsymbol{\Delta}\bm{a}_i-\mathcal{G}_i\boldsymbol{\delta}\bm{f})
    \end{gather}
\end{subequations}
Their corresponding estimate covariances are given by
\begin{equation}
\begin{split}
    P_{b_i}&=\mathcal{E}\{\boldsymbol{\delta}\bm{b}_i\boldsymbol{\delta}\bm{b}_i^T\}\\
    &=\mathcal{E}\{(\hat{\bm{b}}_i-\bm{{b}}_i)(\hat{\bm{b}}_i-\bm{{b}}_i)^T\}\\
    &=R_{b_i}+\text{cov}\{\hat{\bm{b}}_i-\tilde{\bm{b}}_i\}\\
    &+C_iR_{b_i}-C_i\mathcal{G}_i P_{\bm{f}}\mathcal{G}_i^T Q^{-1}_{\lambda_i} R_{b_i}\\
    &+\left(C_iR_{b_i}-C_i\mathcal{G}_i P_{\bm{f}} \mathcal{G}_i^T Q^{-1}_{\lambda_i} R_{b_i}\right)^T\label{eqn:P_bi}
\end{split}
\end{equation}
\begin{equation}
    \begin{split}
        P_{r_i}&=\mathcal{E}\{\boldsymbol{\delta}\bm{r}_i\delta\bm{r}^T_i\}\\
    &=\mathcal{E}\{(\hat{\bm{r}}_i-\bm{{r}}_i)(\hat{\bm{r}}_i-\bm{{r}}_i)^T\}\\
    &=R_{r_i}+\text{cov}\{\hat{\bm{r}}_i-\tilde{\bm{r}}_i\}\\
    &-D_iAR_{r_i}+D_i\mathcal{G}_i P_{\bm{f}}\mathcal{G}_i^T Q^{-1}_{\lambda_i} AR_{r_i}\\
    &+\left(-D_iAR_{r_i}+D_i\mathcal{G}_i P_{\bm{f}}\mathcal{G}_i^T Q^{-1}_{\lambda_i} AR_{r_i}\right)^T\label{eqn:P_ri}
    \end{split}
\end{equation}

\subsection{Isotropic Noise Covariance}
The previous section provided analytical expressions for the covariance of estimates and residuals for the most generic case of observation noise covariance, which is a fully-populated symmetric positive-definite matrix as denoted in Eq.~\eqref{eqn:R_i}.
Now a particular case of previous covariance derivations is elaborated.
In some sensors, there is a simplifying assumption on the distribution of noise having the same characteristics in the different space coordinates of $x$, $y$ and $z$.
This presumption results in a particular form of the noise covariance called \textit{isotropic}.
For the pose estimation problem in this work, this isotropic covariance of observation errors is denoted by
\begin{equation}
R_i=\begin{bmatrix}\sigma^2_{r_i}I_{3\times3} & 0_{3\times3}\\
    0_{3\times3} & \sigma^2_{b_i}I_{3\times3}\end{bmatrix}
\end{equation}
where $\sigma_{r_i}$ and $\sigma_{b_i}$ indicate the standard deviation of noise for the sensors in the reference and body frames, respectively.
The cross-correlation between the reference and body sensor noise will be zero, as shown in the off-diagonal blocks.
For the isotropic case, the weight matrix $Q_{\hat{\lambda}_i}$ in the cost function from Eq.~\eqref{eqn:expand_Q} shrinks to
\begin{equation}
\begin{split}
    Q_{\hat{\lambda}_i}&=\hat{A}R_{r_i}\hat{A}^T+R_{b_i}\\
    &=\hat{A}\sigma^2_{r_i}I_{3\times3}\hat{A}^T+\sigma^2_{b_i}I_{3\times3}\\
    &=\left(\sigma^2_{r_i}+\sigma^2_{b_i}\right)I_{3\times3}\\
    &=\sigma^2_i I_{3\times3}
\end{split}
\end{equation}
where
\begin{equation}
    \sigma_i=\sqrt{\sigma^2_{r_i}+\sigma^2_{b_i}}
\end{equation}
As a result, the second weight $\bar{Q}_{\hat{\lambda}}$ of the cost function in Eq.~\eqref{q_cap_bar} becomes
\begin{subequations}
    \begin{align}
        \bar{Q}_{\hat{\lambda}}&=\bar{\sigma}^{-2}I_{3\times3}\\
        \bar{\sigma}&\equiv\sqrt{\sum_{i=1}^n \sigma^{-2}_i}=\sqrt{\sum_{i=1}^n \frac{1}{\sigma^2_{r_i}+\sigma^2_{b_i}} }
    \end{align}
\end{subequations}
Then the attitude-only cost function in Eq.~\eqref{eqn:cost_ahat} yields
\begin{subequations}
\begin{gather}
    J(\hat{A})=\frac{1}{2}\Bigg[\sum_{i=1}^{n}\sigma^{-2}_i\|\tilde{\bm{b}}_i-\hat{A}\tilde{\bm{r}}_i\|^2\Bigg]-\frac{1}{2}\bar{\sigma}^2\|\Bar{\tilde{\bm{b}}}-\hat{A}\Bar{\tilde{\bm{r}}}\|^2\label{eqn:cost_iso}\\
    \Bar{\tilde{\bm{b}}}=\bar{\sigma}^{-2}\sum_{i=1}^n \sigma^{-2}_i \tilde{\bm{b}}_i\label{eqn:b_bar_iso}\\
    \Bar{\tilde{\bm{r}}}=\bar{\sigma}^{-2}\sum_{i=1}^n \sigma^{-2}_i \tilde{\bm{r}}_i\label{eqn:r_bar_iso}
\end{gather}
\end{subequations}
The first part of the right side of Eq.~\eqref{eqn:cost_iso} is equivalent to Wahba's problem \cite{wahba:65}. The second part of the right side of Eq.~\eqref{eqn:cost_iso} involves a weighted average of the measurements \cite{shuster:00}.

Several closed-form expressions for the attitude estimate that minimizes Eq.~\eqref{eqn:cost_iso} are possible \cite{markley:14}. Here the attitude matrix is itself is estimated. First, the following ``attitude profile matrix'' is defined:
\begin{equation}\label{att_pro_mat}
\begin{split}
  B & \equiv \sum_{i=1}^{n} \sigma^{-2}_i (\tilde{\bm{b}}_i - \bar{\tilde{\bm{b}}})
  (\tilde{\bm{r}}_i - \bar{\tilde{\bm{r}}})^T
  \\
 &  = -\bar{\sigma}^{-2} \bar{\tilde{\bm{b}}} \, \bar{\tilde{\bm{r}}}^T + \sum_{i=1}^{n} \sigma^{-2}_i \tilde{\bm{b}}_i \tilde{\bm{r}}^T_i
\end{split}
\end{equation}
Then, its singular value decomposition is taken:
\begin{equation}\label{b_svd}
  B = U \, \Sigma \, V^T
\end{equation}
The attitude estimate is given by
\begin{equation}\label{svd_ae}
  \hat{A} = U \text{diag}([1 \,\,\, 1 \,\,\, \det(U)\det(V)]) V^T
\end{equation}
where the $\det(U)\det(V)$ term is used to ensure that the estimated attitude matrix is proper. Note that Eq.~\eqref{att_pro_mat} is given by a rank-one perturbed matrix update.

Other solutions using the quaternion parameterization can also be determined using the approaches shown in \cite{markley:14}.

It is noted that in the isotropic case the matrix $Q_{\hat{\lambda}_i}$ in Eq.~\eqref{eqn:cost_ahat} is not a function of attitude matrix. Therefore, the attitude error avoids several complications in the derivation.
This simplicity comes with the cost of ignoring the possible cross-covariances in the reference and body frames and assuming the same statistical properties of noise along all space coordinates.
The cost function in Eq.~\eqref{eqn:cost_iso} is already second-order in terms of the attitude matrix, and there is no need for trimming the higher-order terms, which is another simplicity resulting from the isotropic assumption.

The optimal attitude from Eq.~\eqref{eqn:d_alpha} yields
\begin{equation}
\begin{split}
    \boldsymbol{\delta \alpha}&=\mathcal{H}^{-1}\left[\sum_{j=1}^n \sigma^{-2}_j\mathcal{A}_j\bar{\sigma}^{-2} \sum_{i=1}^n \sigma^{-2}_i\boldsymbol{\Delta}\bm{a}_i-\sum_{i=1}^n\sigma^{-2}_i\mathcal{A}_i\boldsymbol{\Delta}\bm{a}_i\right]\\
    &=\mathcal{H}^{-1}\left[\sum_{i=1}^n \sigma^{-2}_i\bar{\mathcal{A}}\boldsymbol{\Delta}\bm{a}_i-\sum_{i=1}^n\sigma^{-2}_i\mathcal{A}_i\boldsymbol{\Delta}\bm{a}_i\right]\\
    &=\mathcal{H}^{-1}\left[\sum_{i=1}^n \sigma^{-2}_i\left(\bar{\mathcal{A}}-\mathcal{A}_i\right)\boldsymbol{\Delta}\bm{a}_i\right]
\end{split}
\end{equation}
where
\begin{subequations}
\begin{gather}
    \mathcal{A}_i=[A\bm{r}_i\times]\\
    \bar{\mathcal{A}}=\bar{\sigma}^{-2}\sum_{i=1}^n \sigma^{-2}_i[A\bm{r}_i\times]
\end{gather}
\end{subequations}
and the Hessian matrix $\mathcal{H}$ is
\begin{equation}\label{Hess_iso}
    \begin{split}
        \mathcal{H}&=\sum_{i=1}^n\sigma^{-2}_i\mathcal{A}^T_i\mathcal{A}_i-\left(\sum_{i=1}^n\sigma^{-2}_i\mathcal{A}_i\right)^T\bar{\sigma}^{-2}\left(\sum_{i=1}^n\sigma^{-2}_i\mathcal{A}_i\right)\\
        &=-\sum_{i=1}^n\sigma^{-2}_i\mathcal{A}^2_i-\left(\bar{\sigma}^{-2}\sum_{i=1}^n\sigma^{-2}_i\mathcal{A}_i\right)^T\bar{\sigma}^{2}\left(\bar{\sigma}^{-2}\sum_{i=1}^n\sigma^{-2}_i\mathcal{A}_i\right)\\
        &=-\sum_{i=1}^n\sigma^{-2}_i\mathcal{A}^2_i+\bar{\sigma}^{2}\bar{\mathcal{A}}^2
    \end{split}
\end{equation}
The skew-symmetric property of matrices $\mathcal{A}_i$ and $\bar{\mathcal{A}}$ is employed in the above derivation, which originates from the skew-symmetric nature of the cross product matrices.

Now the attitude error-covariance can be computed.
From Eq.~\eqref{eqn:P_delta_alpha_final} and Eq.~\eqref{Hess_iso}, this becomes
\begin{equation}
    P_{\boldsymbol{\delta\alpha}}=\left[-\sum_{i=1}^n\sigma^{-2}_i\mathcal{A}^2_i+\bar{\sigma}^{2}\bar{\mathcal{A}}^2\right]^{-1}
\end{equation}
Regarding the position estimates, the estimate errors are derived from Eq.~\eqref{eqn:err_p} as
\begin{equation}
    \begin{split}
        \boldsymbol{\delta} \bm{p}&=-\bar{\sigma}^{-2}\sum_{i=1}^{n}\sigma^{-2}_i\left(\boldsymbol{\Delta} \bm{a}_i-\mathcal{A}_i\boldsymbol{\delta} \boldsymbol{\alpha}\right) \\
    &=-\boldsymbol{\Delta}\bar{\bm{a}}+\bar{\mathcal{A}}\boldsymbol{\delta\alpha}\\
    \end{split}
\end{equation}
where
\begin{equation}
    \begin{split}
      \boldsymbol{\Delta}\bar{\bm{a}}&\equiv\bar{\sigma}^{-2}\sum_{i=1}^n\sigma^{-2}_i\boldsymbol{\Delta}\bm{a}_i
    \end{split}
\end{equation}    
The resulting error-covariance is obtained from Eq.~\eqref{calc_cov_phat}.
For the isotropic case, this will be
\begin{equation}
    \text{cov}\{\hat{\bm{p}}\}=\bar{\sigma}^{-2}I_{3\times3}-\bar{\mathcal{A}}\left[-\sum_{i=1}^n\sigma^{-2}_i\mathcal{A}^2_i+\bar{\sigma}^{2}\bar{\mathcal{A}}^2\right]^{-1}\bar{\mathcal{A}}
\end{equation}
The observation vector estimates can be computed from Eq.~\eqref{eqn:bhat-btilde} and Eq.~\eqref{eqn:rhat-rtilde}, and for the isotropic covariance will result in
\begin{subequations}
\begin{gather}
    \hat{\bm{b}}_i=\tilde{\bm{b}}_i-\frac{\sigma^2_{b_i}}{\sigma^2_{b_i}+\sigma^2_{r_i}}(\tilde{\bm{b}}_i-\hat{A}\tilde{\bm{r}}_i+\hat{\bm{p}})\\
    \hat{\bm{r}}_i=\tilde{\bm{r}}_i+\frac{\sigma^2_{r_i}}{\sigma^2_{r_i}+\sigma^2_{b_i}}\hat{A}^T(\tilde{\bm{b}}_i-\hat{A}\tilde{\bm{r}}_i+\hat{\bm{p}})
\end{gather}
\end{subequations}
From the first-order approximation of observation residuals in Eqs.~\eqref{apprxb-btilde} and \eqref{apprxr-rtilde}, the observation residuals are given as
\begin{subequations}
\begin{gather}
    \boldsymbol{\delta}\bm{b}_i-\boldsymbol{\Delta}\bm{b}_i\approx -\frac{\sigma^2_{b_i}}{\sigma^2_{b_i}+\sigma^2_{r_i}}(\boldsymbol{\Delta}\bm{a}_i-\mathcal{G}_i\boldsymbol{\delta}\bm{f})\\
    \boldsymbol{\delta}\bm{r}_i-\boldsymbol{\Delta}\bm{r}_i\approx \frac{\sigma^2_{r_i}}{\sigma^2_{b_i}+\sigma^2_{r_i}}A^T(\boldsymbol{\Delta}\bm{a}_i-\mathcal{G}_i\boldsymbol{\delta}\bm{f})
\end{gather}
\end{subequations}
with the definitions of $\boldsymbol{\Delta}\bm{a}_i$, $\mathcal{G}_i$, and $\boldsymbol{\delta}\bm{f}$ in Eqs.~\eqref{eqn:def_Dai}, \eqref{eqn:def_Gi} and \eqref{eqn:def_delta_f}, respectively.
Then, the observation residual covariances become
\begin{subequations}
\begin{align}
    \text{cov}(\hat{\bm{b}}_i-\tilde{\bm{b}}_i)&=\frac{\sigma^2_{b_i}}{\left(\sigma^2_{b_i}+\sigma^2_{r_i}\right)^2} \left[\left(\sigma^2_{b_i}+\sigma^2_{r_i}\right)I_{3\times3}-\begin{bmatrix}\mathcal{A}_i&-I_{3\times3}\end{bmatrix} \left(\sum_{i=1}^n\frac{1}{\sigma^2_{b_i}+\sigma^2_{r_i}}\begin{bmatrix}-\mathcal{A}^2_i&\mathcal{A}_i\\\mathcal{A}^T_i&I_{3\times3}\end{bmatrix}\right)^{-1} \begin{bmatrix}\mathcal{A}^T_i\\-I_{3\times3}\end{bmatrix}\right]\\
    \text{cov}(\hat{\bm{r}}_i-\tilde{\bm{r}}_i)&=\frac{\sigma^2_{r_i}}{\left(\sigma^2_{b_i}+\sigma^2_{r_i}\right)^2}A^T \left[\left(\sigma^2_{b_i}+\sigma^2_{r_i}\right)I_{3\times3}-\begin{bmatrix}\mathcal{A}_i&-I_{3\times3}\end{bmatrix} \left(\sum_{i=1}^n\frac{1}{\sigma^2_{b_i}+\sigma^2_{r_i}}\begin{bmatrix}-\mathcal{A}^2_i&\mathcal{A}_i\\\mathcal{A}^T_i&I_{3\times3}\end{bmatrix}\right)^{-1} \begin{bmatrix}\mathcal{A}^T_i\\-I_{3\times3}\end{bmatrix}\right] A
\end{align}
\end{subequations}
The estimate covariances of the observation vectors are simplified from Eqs.~\eqref{eqn:P_bi} and \eqref{eqn:P_ri} as
\begin{equation}
    \begin{split}
        P_{b_i}&=\sigma^2_{b_i}I_{3\times3}+\text{cov}\{\hat{\bm{b}}_i-\tilde{\bm{b}}_i\}\\
    &-2\frac{\sigma^3_{b_i}}{\sigma^2_{b_i}+\sigma^2_{r_i}}I_{3\times3}+\frac{\sigma^3_{b_i}}{\left(\sigma^2_{b_i}+\sigma^2_{r_i}\right)^2}\begin{bmatrix}\mathcal{A}_i&-I_{3\times3}\end{bmatrix} \left(\sum_{i=1}^n\frac{1}{\sigma^2_{b_i}+\sigma^2_{r_i}}\begin{bmatrix}-\mathcal{A}^2_i&\mathcal{A}_i\\\mathcal{A}^T_i&I_{3\times3}\end{bmatrix}\right)^{-1} \begin{bmatrix}\mathcal{A}^T_i\\-I_{3\times3}\end{bmatrix}\\
    &+\frac{\sigma^3_{b_i}}{\left(\sigma^2_{b_i}+\sigma^2_{r_i}\right)^2}\begin{bmatrix}\mathcal{A}_i&-I_{3\times3}\end{bmatrix} \left(\sum_{i=1}^n\frac{1}{\sigma^2_{b_i}+\sigma^2_{r_i}}\begin{bmatrix}-\mathcal{A}^2_i&\mathcal{A}_i\\\mathcal{A}^T_i&I_{3\times3}\end{bmatrix}\right)^{-T} \begin{bmatrix}\mathcal{A}^T_i\\-I_{3\times3}\end{bmatrix}
    \end{split}
\end{equation}
\begin{equation}
    \begin{split}
        P_{r_i}&=\sigma^2_{r_i}I_{3\times3}+\text{cov}\{\hat{\bm{r}}_i-\tilde{\bm{r}}_i\} \\
    &-2\frac{\sigma^3_{r_i}}{\sigma^2_{b_i}+\sigma^2_{r_i}}I_{3\times3}+\frac{\sigma^3_{r_i}}{\left(\sigma^2_{b_i}+\sigma^2_{r_i}\right)^2}A^T\begin{bmatrix}\mathcal{A}_i&-I_{3\times3}\end{bmatrix} \left(\sum_{i=1}^n\frac{1}{\sigma^2_{b_i}+\sigma^2_{r_i}}\begin{bmatrix}-\mathcal{A}^2_i&\mathcal{A}_i\\\mathcal{A}^T_i&I_{3\times3}\end{bmatrix}\right)^{-1} \begin{bmatrix}\mathcal{A}^T_i\\-I_{3\times3}\end{bmatrix}A\\
    &+\frac{\sigma^3_{b_i}}{\left(\sigma^2_{b_i}+\sigma^2_{r_i}\right)^2}A^T\begin{bmatrix}\mathcal{A}_i&-I_{3\times3}\end{bmatrix} \left(\sum_{i=1}^n\frac{1}{\sigma^2_{b_i}+\sigma^2_{r_i}}\begin{bmatrix}-\mathcal{A}^2_i&\mathcal{A}_i\\\mathcal{A}^T_i&I_{3\times3}\end{bmatrix}\right)^{-T} \begin{bmatrix}\mathcal{A}^T_i\\-I_{3\times3}\end{bmatrix}A
    \end{split}
\end{equation}
All of the covariance expressions reduce down to the ones derived in \cite{cheng2021optimal}.

It is well known that at least two nonparallel vector observations must exist in order for the attitude to be observable in Wahba's standard problem \cite{wahba:65}. It is now shown that at least three nonparallel vectors must exist for the attitude to be observable in the current pose estimation problem. For $n=2$,  Eqs.~\eqref{eqn:r_bar_iso} and \eqref{eqn:b_bar_iso} give
\begin{subequations}
\begin{gather}
\bar{\tilde{\bm{r}}} = \frac{\sigma_{2}^{2} \tilde{\bm{r}}_{1} + \sigma_{1}^{2} \tilde{\bm{r}}_{2}}{\sigma_{1}^{2} + \sigma_{2}^{2}}
\\
\bar{\tilde{\bm{b}}} = \frac{\sigma_{2}^{2} \tilde{\bm{b}}_{1} + \sigma_{1}^{2} \tilde{\bm{b}}_{2}}{\sigma_{1}^{2} + \sigma_{2}^{2}}
\end{gather}
\end{subequations}
Then,
\begin{subequations}
\begin{gather}
\tilde{\bm{r}}_{1} - \bar{\tilde{\bm{r}}}  = \frac{\sigma_{1}^{2}}{\sigma_{1}^{2}+\sigma_{2}^{2}}(\tilde{\bm{r}}_{1}-\tilde{\bm{r}}_{2})
\\
\tilde{\bm{r}}_{2} - \bar{\tilde{\bm{r}}}  = -\frac{\sigma_{2}^{2}}{\sigma_{2}^{2}+\sigma_{2}^{2}}
(\tilde{\bm{r}}_{1}-\tilde{\bm{r}}_{2})
\\
\tilde{\bm{b}}_{1} - \bar{\tilde{\bm{b}}}  = \frac{\sigma_{1}^{2}}{\sigma_{1}^{2}+\sigma_{2}^{2}}(\tilde{\bm{b}}_{1}-\tilde{\bm{b}}_{2})
\\
\tilde{\bm{b}}_{2} - \bar{\tilde{\bm{b}}}  = -\frac{\sigma_{2}^{2}}{\sigma_{2}^{2}+\sigma_{2}^{2}}
(\tilde{\bm{b}}_{1}-\tilde{\bm{b}}_{2})
\end{gather}
\end{subequations}
Thus, the matrix $B$ in Eq.~\eqref{att_pro_mat} is explicitly given by
\begin{equation}\label{b2_observ}
\begin{split}
  B &= \frac{\sigma_{1}^{2}}{(\sigma_{1}^{2} + \sigma_{2}^{2})^{2}}(\tilde{\bm{b}}_1 -\tilde{\bm{b}}_2) (\tilde{\bm{r}}_1 -\tilde{\bm{r}}_2)^T + \frac{\sigma_{2}^{2}}{(\sigma_{1}^{2} + \sigma_{2}^{2})^{2}}(\tilde{\bm{b}}_1 -\tilde{\bm{b}}_2) (\tilde{\bm{r}}_1 -\tilde{\bm{r}}_2)^T
  \\
 &= \frac{1}{\sigma_{1}^{2} + \sigma_{2}^{2}}(\tilde{\bm{b}}_1 -\tilde{\bm{b}}_2) (\tilde{\bm{r}}_1 -\tilde{\bm{r}}_2)^T
  \end{split}
\end{equation}
The matrix $B$ must at least have rank 2 for the attitude to be observable \cite{markley:14}. Clearly, for the two-vector case the matrix $B$ in Eq.~\eqref{b2_observ} has rank 1, which means that for this problem the attitude is not observable. This intuitively makes sense. There are six unknowns for this problem: three for the attitude and three for the position. Each vector observation provides two pieces of information (rotations around the vector are unknown). Thus, intuitively, at least three nonparallel vectors must exist to fully solve this pose estimation problem.

Similarly, for $n=2$, the matrix $\mathcal{H}$ (the FIM) is singular and thus the attitude-error covariance is not well defined.

Since
\begin{subequations}
\begin{gather}
\bm{r}_{1} - \Bar{\tilde{\bm{r}}} = \frac{\sigma_{1}^{2}}{\sigma_{1}^{2} + \sigma_{2}^{2}}(\bm{r}_{1}-\bm{r}_{2})\\
\bm{r}_{2} - \Bar{\tilde{\bm{r}}} = -\frac{\sigma_{2}^{2}}{\sigma_{1}^{2} + \sigma_{2}^{2}}(\bm{r}_{1}-\bm{r}_{2})
\end{gather}
\end{subequations}
then
\begin{equation}
\mathcal{H} = -\frac{\sigma_{1}^{2}}{(\sigma_{1}^{2}+\sigma_{2}^{2})^{2}}[A(\bm{r}_{1}-\bm{r}_{2})\times]^{2} -\frac{\sigma_{2}^{2}}{(\sigma_{1}^{2}+\sigma_{2}^{2})^{2}}[A(\bm{r}_{1}-\bm{r}_{2})\times]^{2}  = - \frac{1}{\sigma_{1}^{2}+\sigma_{2}^{2}} [A(\bm{r}_{1}-\bm{r}_{2})\times]^{2}
\end{equation}
Clearly, $\mathcal{H} A(\bm{r}_{1}-\bm{r}_{2}) = \bm{0}_{3\times 1}$. Hence, if $n=2$, then $\mathcal{H}$ is singular and the null vector of $\mathcal{H}$ is $A(\bm{r}_{1}-\bm{r}_{2})$.
The $n$ vectors $\bm{r}_i-\bar{\bm{r}}$ are always linearly dependent, but if $n=2$, then the two dependent vectors are antiparallel to each other, which makes $\mathcal{H}$ singular.

\subsection{Fisher Information Matrix}
From the analysis in Section \ref{cov_analysis}, the estimate error-covariances for the attitude error $\boldsymbol{\delta} \boldsymbol{\alpha}$, the translation vector $\bm{p}$, the residuals as well as the estimate covariances for the observation vectors $\tilde{\bm{b}}_i$ and $\tilde{\bm{r}}_i$ have been derived. An efficiency proof for the estimate error-covariances of the attitude and translation vector is now shown based on the FIM and the CRLB defined for the estimation covariances. For an unbiased estimate $\hat{\bm{f}}$, the the following inequality exists \cite{crassidis:12}:
\begin{equation}
    \text{cov}\{\hat{\bm{f}}\}\geq F^{-1}\equiv \left[\mathcal{E}\left\{-\frac{\partial^2}{\partial \hat{\bm{f}} \partial \hat{\bm{f}}^T}p(\bm{\tilde{y}|\hat{\bm{f}}})\right\}\right]^{-1}\label{eqn:crlb}
\end{equation}
The term inside of the expectation shows the Hessian of the the negative log-likelihood function, which is given in Eq.~\eqref{eqn:def_sec_order_cost}.
For an optimal estimator, the equality in the Eq.~\eqref{eqn:crlb} is given, and the estimator is {\it efficient}. The Hessian of the second-order approximated cost function $J$ is the FIM. From Eqs.~\eqref{bti-ati} and Eq.~\eqref{phat-p}, the following are given:
\begin{equation}
\begin{split}
    \tilde{\bm{b}}_i-\hat{A}\tilde{\bm{r}}_i+\hat{\bm{p}}&\approx-\bm{p}+\boldsymbol{\Delta}\bm{a}_i-\mathcal{A}_i\boldsymbol{\delta}\boldsymbol{\alpha}+\bm{p}+\boldsymbol{\delta} \bm{p}\\
    &=\boldsymbol{\Delta} \bm{a}_i-\mathcal{A}_i\boldsymbol{\delta}\boldsymbol{\alpha}+\boldsymbol{\delta} \bm{p} \label{del_brp}
\end{split}
\end{equation}
The second-order cost function involving $\boldsymbol{\delta}{\bm{p}}$ is now given by
\begin{equation}
    J =\frac{1}{2}\sum_{i=1}^n\left(\boldsymbol{\Delta} \bm{a}_i-\mathcal{A}_i\boldsymbol{\delta}\boldsymbol{\alpha}+\boldsymbol{\delta} \bm{p}\right)^T Q^{-1}_{\lambda_i}\left(\boldsymbol{\Delta} \bm{a}_i-\mathcal{A}_i\boldsymbol{\delta}\boldsymbol{\alpha}+\boldsymbol{\delta} \bm{p}\right)
\end{equation}
The FIM, denoted by $F$, will be
\begin{align}
    F=\begin{bmatrix} \sum_{i=1}^{n}{\mathcal{A}_i^T Q_{\lambda_i}^{-1}\mathcal{A}_i} & -\sum_{i=1}^{n}{\mathcal{A}_i^T Q_{\lambda_i}^{-1}}\\
     -\sum_{i=1}^{n}{Q_{\lambda_i}^{-1}\mathcal{A}_i}& \sum_{i=1}^{n}Q_{\lambda_i}^{-1}\end{bmatrix}
\end{align}
The block matrices of the FIM are
\begin{align}
    F=\begin{bmatrix} F_{11} & F_{12}\\
     F_{21}& F_{22}\end{bmatrix}
\end{align}
The inverse of FIM, denoted by $\mathcal{F}$, is given by
\begin{align}
    \mathcal{F}=\begin{bmatrix} \mathcal{F}_{11} & \mathcal{F}_{12}\\
    \mathcal{F}_{21}& \mathcal{F}_{22}\end{bmatrix}
\end{align}
The $\mathcal{F}_{11}$ block of this matrix is calculated by the Sherman–Morrison–Woodbury lemma \cite{sherman1950adjustment}:
\begin{equation}
\begin{split}
    \mathcal{F}_{11}&=\left(F_{11}-F_{12}F^{-1}_{22}F_{21}\right)^{-1}\\
    &=\left[\sum_{i=1}^{n}{\mathcal{A}_i^T Q_{\lambda_i}^{-1}\mathcal{A}_i}-\left(-\sum_{i=1}^{n}{\mathcal{A}_i^T Q_{\lambda_i}^{-1}}\right)S_{\lambda}\left(-\sum_{i=1}^{n}{Q_{\lambda_i}^{-1}\mathcal{A}_i}\right)\right]^{-1}\\
    &=\mathcal{H}^{-1}
\end{split}
\end{equation}
This has already been proven in Eq.~\eqref{eqn:P_delta_alpha_final}, which is the CRLB for the attitude error.
The term $\mathcal{F}_{22}$ is also derived by using matrix inversion lemma as
\begin{equation}
\begin{split}
    \mathcal{F}_{22}&=\left(F_{22}-F_{21}F^{-1}_{11}F_{12}\right)^{-1}\\
    &=\left[\left(\sum_{i=1}^{n}Q_{\lambda_i}^{-1}\right)-\left(-\sum_{i=1}^{n}Q_{\lambda_i}^{-1}\mathcal{A}_i\right)\left(\sum_{i=1}^{n}\mathcal{A}_i^T Q_{\lambda_i}^{-1}\mathcal{A}_i\right)^{-1}
    \left(-\sum_{i=1}^{n}\mathcal{A}_i^T Q_{\lambda_i}^{-1}\right)\right]^{-1}\\
    &=\left[S^{-1}_{\lambda}-S^{-1}_{\lambda}\bar{\mathcal{A}}\left(\sum_{i=1}^{n}\mathcal{A}_i^T Q_{\lambda_i}^{-1}\mathcal{A}_i\right)^{-1}\bar{\mathcal{A}}^TS^{-1}_{\lambda}\right]^{-1}
\end{split}
\end{equation}
Using Eq.~\eqref{eqn:P_delta_alpha_final} leads to
\begin{equation}
\begin{split}
    \mathcal{F}_{22}&=S_{\lambda}+\bar{\mathcal{A}}\left[\sum_{i=1}^{n}\mathcal{A}_i^T Q_{\lambda_i}^{-1}\mathcal{A}_i-\sum_{i=1}^{n}\mathcal{A}_i^T Q_{\lambda_i}^{-1} S_{\lambda}\sum_{i=1}^{n} Q_{\lambda_i}^{-1}\mathcal{A}_i\right]^{-1}\bar{\mathcal{A}}^T\\
    &=S_{\lambda}+\bar{\mathcal{A}}\text{cov}\{\boldsymbol{\delta}\boldsymbol{\alpha}\}\bar{\mathcal{A}}^T
\end{split}
\end{equation}
This proves the CRLB  for covariance of translation vector estimate. Note that from Eq.~\eqref{eqn:P_delta_alpha_final} it can be concluded that the CRLB holds for the attitude estimate because the estimate error-covariance is equal to the inverse of the Hessian of the cost function in Eq.~\eqref{eqn:def_sec_order_cost}.

\section{Simulated and Experimental Validations}\label{numerical_validation}
This section provides a simulated and experimental validations of the TLS solution for pose estimation, as well as the associated covariance expressions. 
For the simulated validation, Monte Carlo simulations are introduced and employed to perform several realizations of the noise in the observation data, and interpret the results of the estimation in the sense of comparing estimation errors and estimation 3$\sigma$ bounds, derived from the estimate error-covariances. Also, references \cite{maleki2021total} and \cite{maleki2022thesis} elaborate on this simulation and showcases that the covariance expressions are accurate through Monte Carlo simulations. The experimental validation is done using a LIDAR system with an accurate ground truth.

\begin{figure}[ht!]
  \begin{centering}
      \subfigure[{\bf Attitude Errors}]
      {\includegraphics[width=0.49\textwidth]{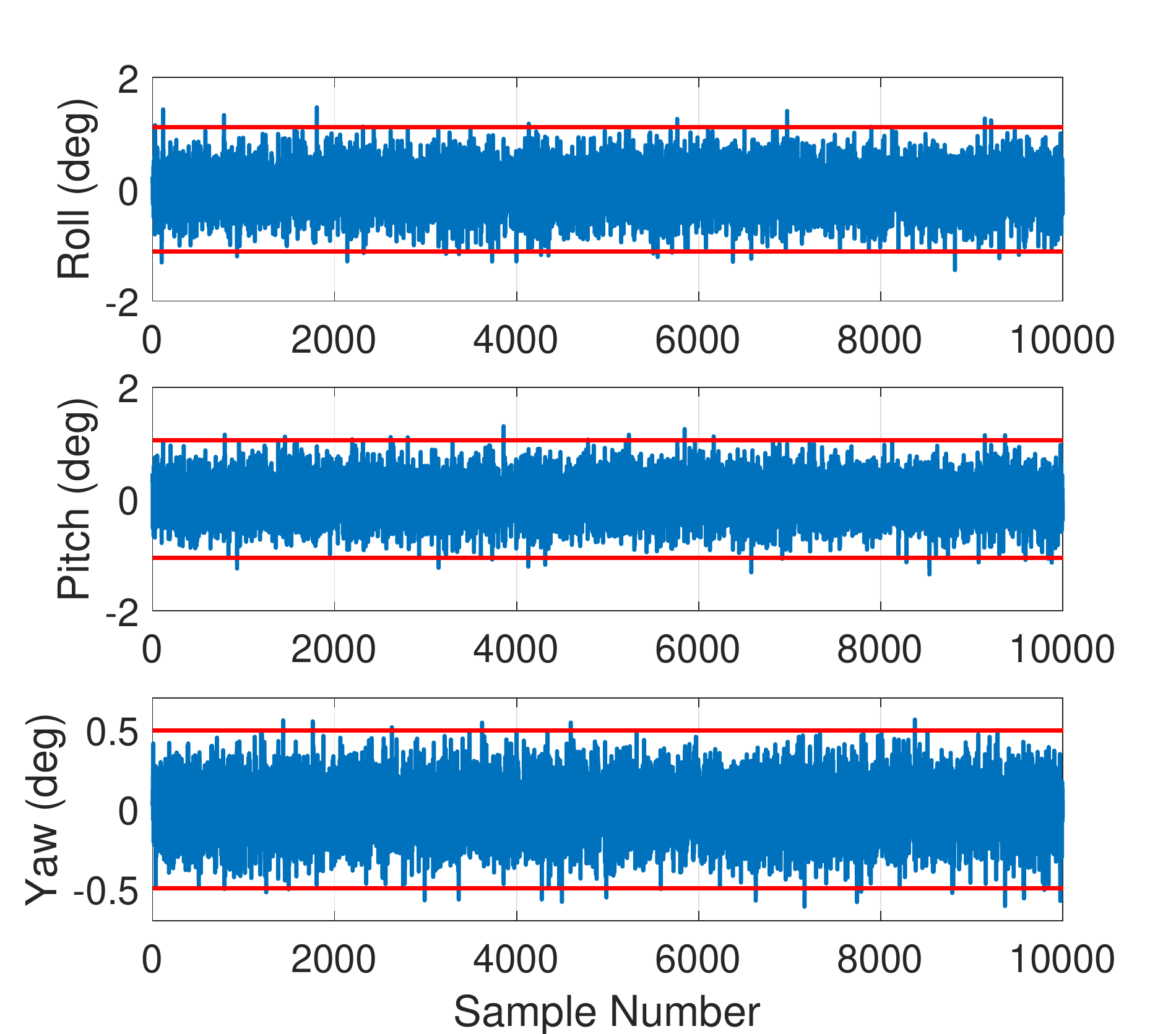}\label{fig:rpy}}
            \subfigure[{\bf Position Errors}]
      {\includegraphics[width=0.49\textwidth]{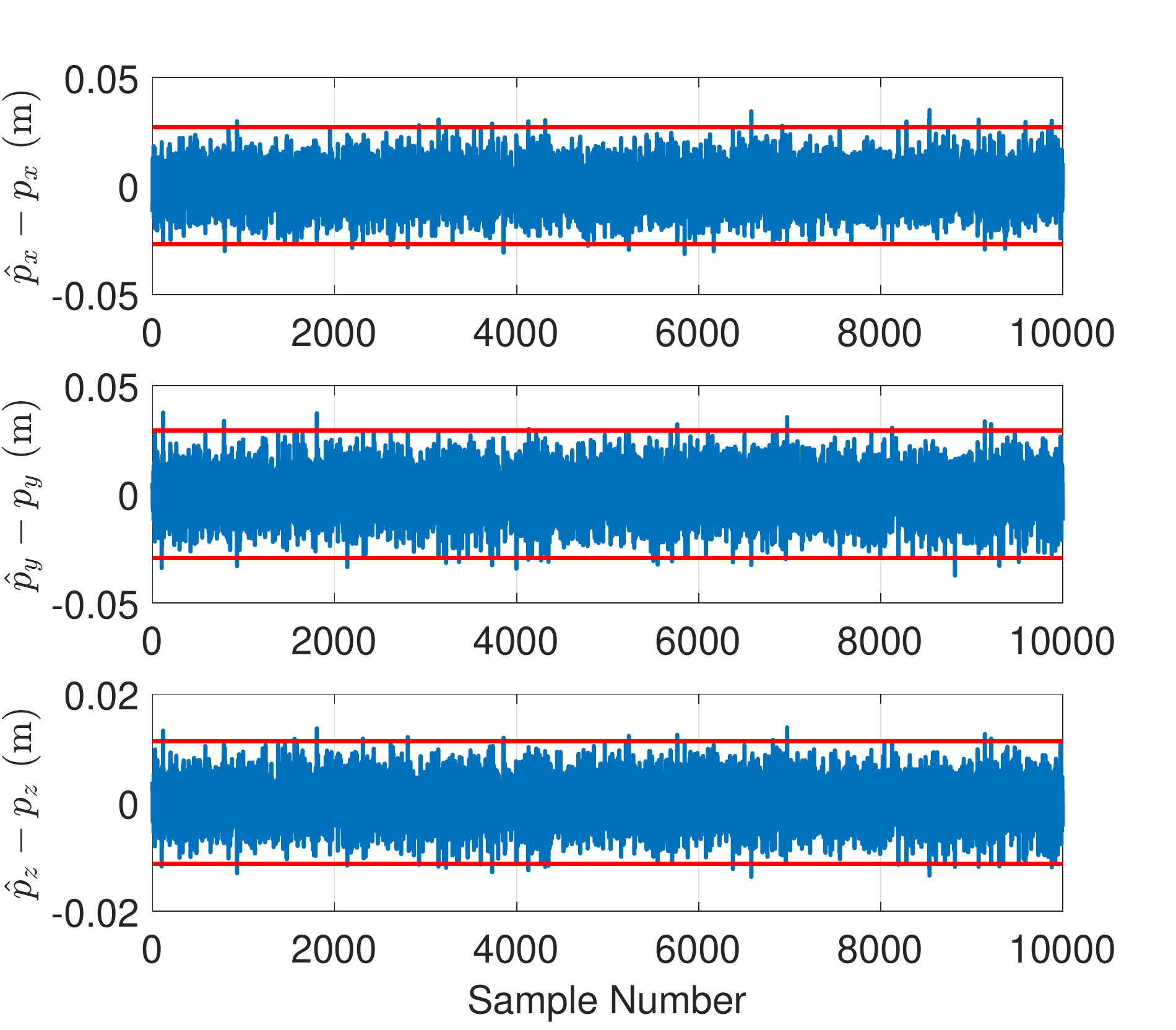}\label{fig:p}}
    \caption{Monte Carlo simulation for the attitude and translation vector.}
  \end{centering}
\end{figure}

\begin{figure}[ht!]
  \begin{centering}
      \subfigure[{\bf Estimation Errors}]
      {\includegraphics[width=0.49\textwidth]{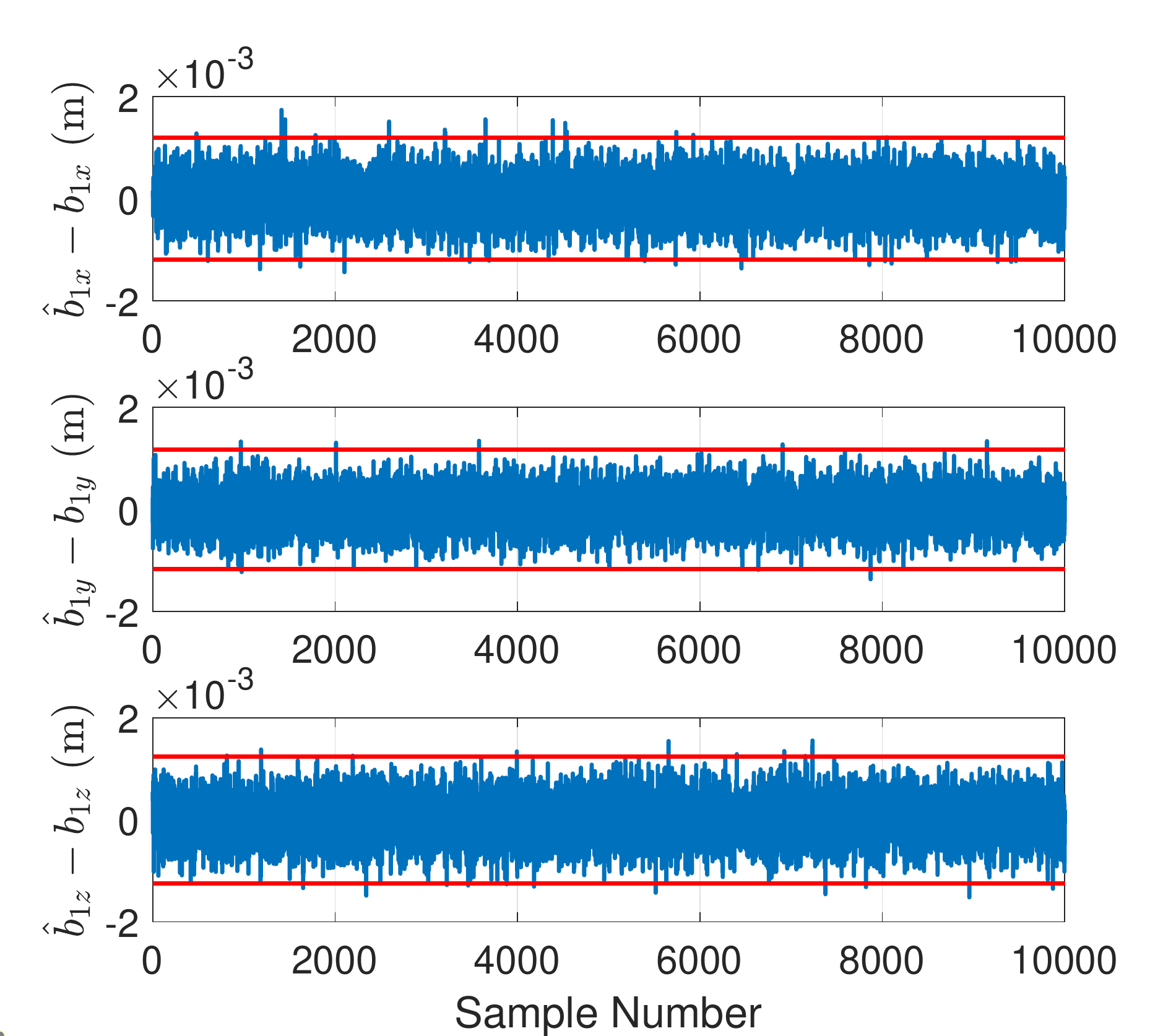}\label{fig:b_est}}
            \subfigure[{\bf Residual Errors}]
      {\includegraphics[width=0.49\textwidth]{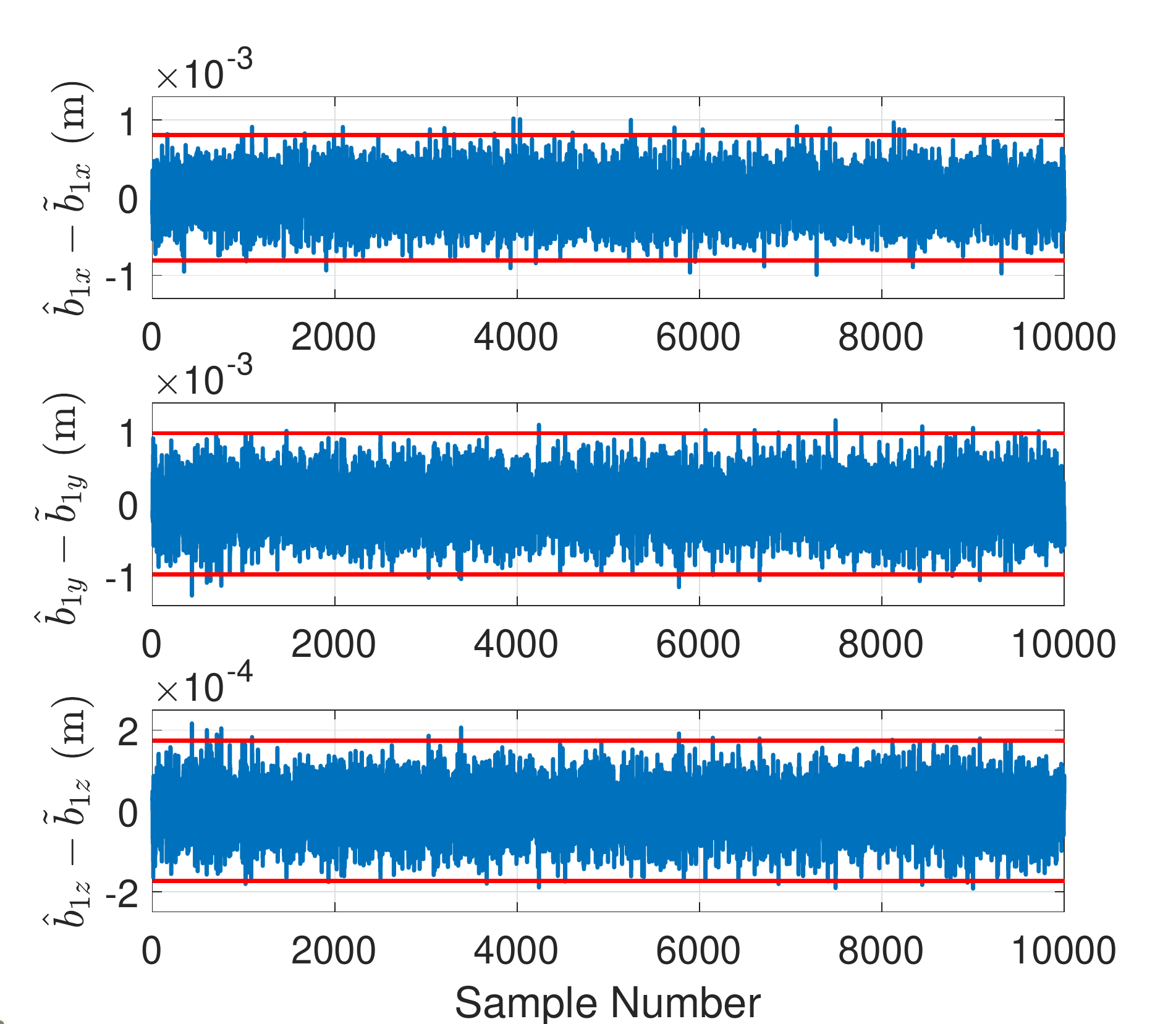}\label{fig:b_res}}
    \caption{Monte Carlo simulation for estimates and residuals of the vector observation $\bm{b}_1$.}
  \end{centering}
\end{figure}

\begin{figure}[ht!]
  \begin{centering}
      \subfigure[{\bf Estimation Errors}]
      {\includegraphics[width=0.49\textwidth]{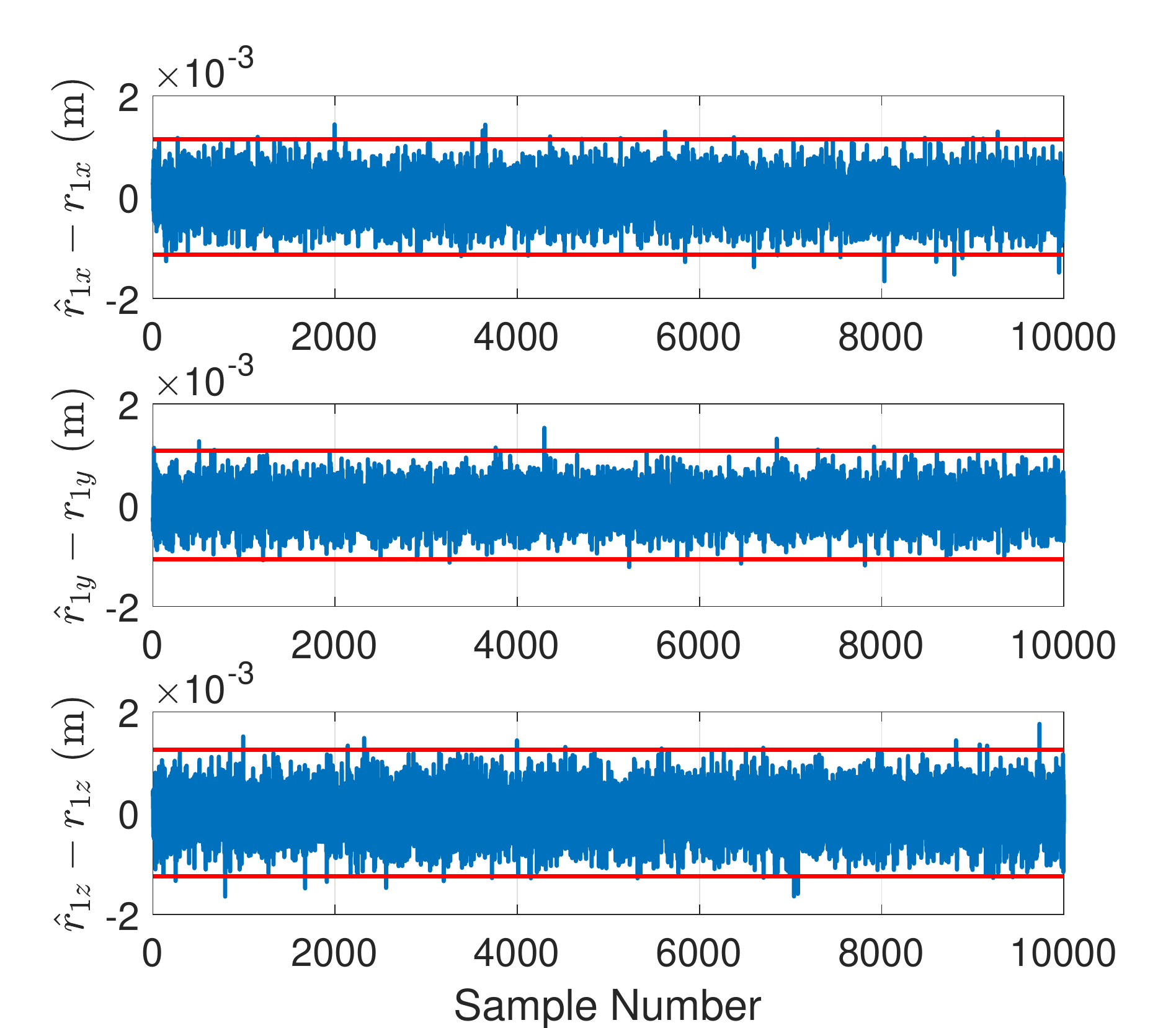}\label{fig:r_est}}
            \subfigure[{\bf Residual Errors}]
      {\includegraphics[width=0.49\textwidth]{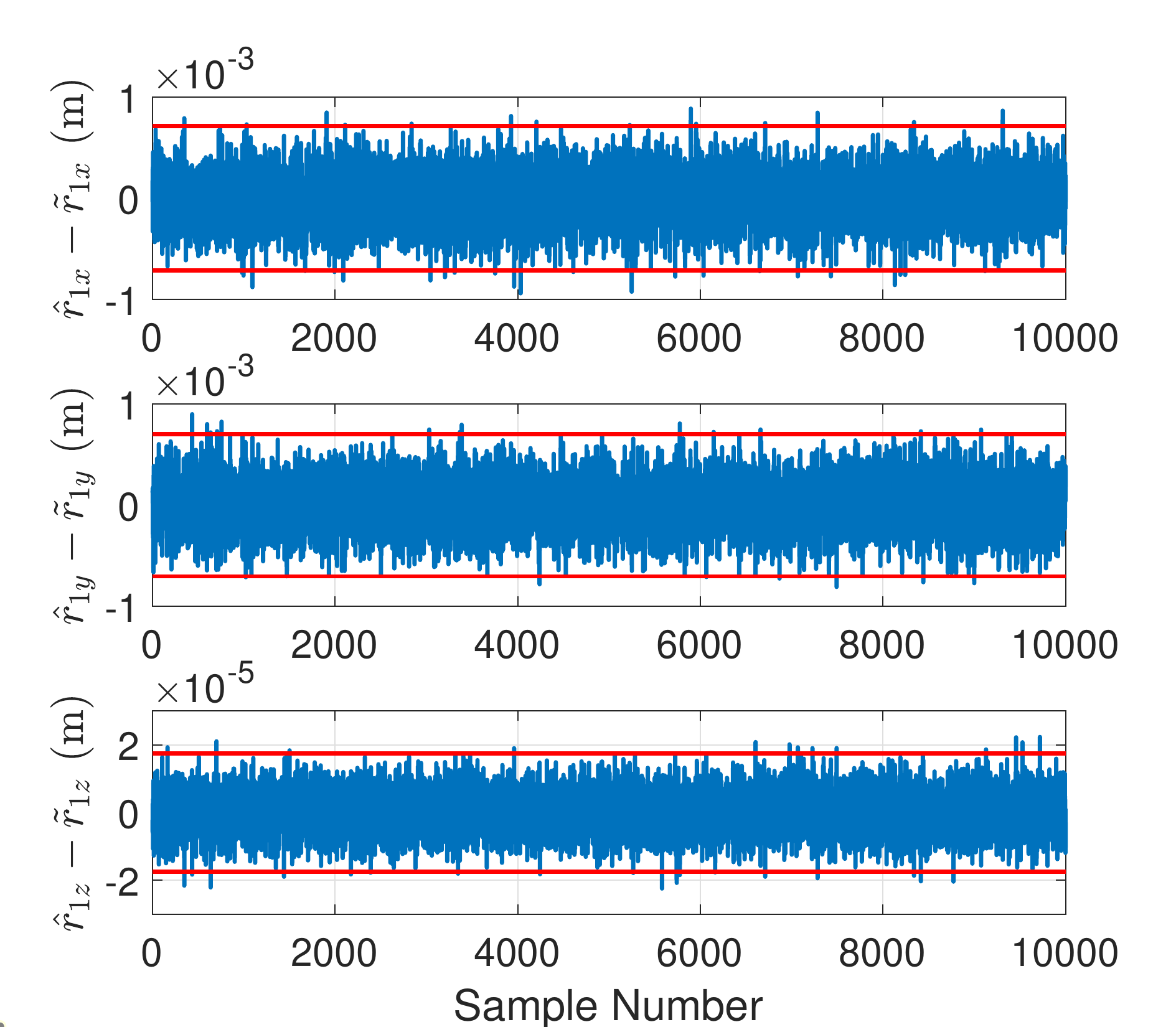}\label{fig:r_res}}
    \caption{Monte Carlo simulation for estimates and residuals of the vector observation $\bm{r}_1$.}
  \end{centering}
\end{figure}

\subsection{Monte Carlo Simulations}
A pose estimation problem is solved here with two scans of a vector-observation-enabled sensor, which can be a LIDAR, with three vector observations per scan. The ground truth values for the attitude matrix $A$, translation vector $\bm{p}$ and vector observations $\bm{b}_i$, $i=1,\,2,\,3$, as well as the measurement covariances, are given as
\begin{gather*}
    A=I_{3\times3}\\
    \bm{p}=\begin{bmatrix}0.3 & -0.4 & 0.5\end{bmatrix}^T \quad \text{m}\\
    \bm{b}_1=\begin{bmatrix} 0 & 9.7590\times 10^{-2} & -1.4833 \times10^{-1}\end{bmatrix}^T \quad \text{m} \\
    \bm{b}_2=\begin{bmatrix} 0 & 1.9518\times 10^{-1} & -1.2855 \times10^{-2}\end{bmatrix}^T \quad \text{m} \\
    \bm{b}_3=\begin{bmatrix} 1 & 9.7590\times 10^{-1} & 9.8885 \times10^{-1}\end{bmatrix}^T\quad \text{m} \\
    R_1=\begin{bmatrix}0.1902 & 0.0228&   -0.0190&   -0.0345&   -0.0079&    0.0225\\
    0.0228&    0.2288&   -0.0003&    0.0145&    0.0483&   -0.0161\\
   -0.0190&   -0.0003&    0.3554&    0.0765&   -0.0180&    0.1386\\
   -0.0345&    0.0145&    0.0765&    0.2566&   -0.0201&    0.0408\\
   -0.0079&    0.0483&   -0.0180&   -0.0201 &   0.2621&   -0.0800\\
    0.0225&   -0.0161&    0.1386&    0.0408&   -0.0800&    0.3349\end{bmatrix} \times 10^{-6} \quad \text{m}^2 \\
    R_2=\begin{bmatrix}0.1981&    0.0213&    0.0021&   -0.0519&   -0.0218&   -0.0231\\
    0.0213&    0.1980&   -0.0264&    0.0023&   -0.0116&    0.0030\\
    0.0021&   -0.0264&    0.2040&   -0.0456 &   0.0273&   -0.0152\\
   -0.0519 &   0.0023 &  -0.0456&    0.2481&    0.0025&    0.0258\\
   -0.0218&   -0.0116&    0.0273&    0.0025&    0.1933&    0.0069\\
   -0.0231&    0.0030&   -0.0152&    0.0258&    0.0069&    0.1851\end{bmatrix} \times 10^{-6}\quad \text{m}^2 \\
   R_3=\begin{bmatrix}0.1705&   -0.0071&   -0.0154&   -0.0247&    0.0081&    0.0049\\
   -0.0071&    0.2036&    0.0038&    0.0259&   -0.0311&    0.0064\\
   -0.0154 &   0.0038 &   0.1910&    0.0376&    0.0085&    0.0166\\
   -0.0247&    0.0259&    0.0376&    0.2738&   -0.0153&    0.0170\\
    0.0081 &  -0.0311 &   0.0085&   -0.0153&    0.1850&   -0.0114\\
    0.0049&    0.0064&    0.0166&    0.0170&   -0.0114&    0.2049\end{bmatrix} \times 10^{-6} \quad \text{m}^2
\end{gather*}
The true observation vectors $\bm{r}_i$ are generated by the constraint in Eq.~\eqref{eqn:det_tls_sensor_model}.
Both measurement and actual observation vectors have a meter unit.
A Monte Carlo simulation with 10,000 samples is performed here to showcase how well the $3\sigma$ bounds generated by the covariance expressions in Eqs.~\eqref{eqn:P_delta_alpha_final} and \eqref{calc_cov_phat} for the attitude and translation vectors, respectively, Eqs.~\eqref{eqn:P_bi} and \eqref{eqn:P_ri} for estimated observations, and Eqs.~\eqref{eqn:cov_bi} and \eqref{eqn:cov_ri} for residuals of observations are bounding their corresponding estimate errors or residuals.
Artificial noise is generated from a Gaussian distribution with zero mean and covariance of $R_{r_i}$, $R_{b_i}$ and $R_{rb_i}$ to produce $\tilde{\bm{r}}_i$ and $\tilde{\bm{b}}_i$ samples.
The covariance matrices of the measurements $R_{r_i}$, $R_{b_i}$ and $R_{rb_i}$ are generated randomly with a signal-to-noise ratio of around $10^{-4}$. Note that by default, the measurement covariance is selected to be positive-definite while being random.
Singularities in the measurement covariance matrix can be handled if they exist, but this is not the focus of this paper.

Figure \ref{fig:rpy} shows the attitude errors in terms of roll, pitch and yaw angles in degrees from the Monte Carlo samples. The blue line depicts the estimation errors, and the red lines are the $3\sigma$ bounds computed from the estimate error-covariances. Figure \ref{fig:p} shows the translation vector estimate error in the $x$, $y$ and $z$ directions, respectively.
It is seen that the estimate errors are well-bounded by their corresponding $3\sigma$ bounds. Figure \ref{fig:b_est} shows the estimation errors, and Fig.~\ref{fig:b_res} shows the residual errors, for observation vector $\bm{b}_1$.
Figures \ref{fig:r_est} and \ref{fig:r_res} show the same results for the observation vector $\bm{r}_1$, respectively.
It is seen that observation vectors are also bounded by their corresponding $3\sigma$ bounds, provided by the covariances of estimates as well as the residuals. 

\subsection{Experimental Results}\label{ch7:experimental}
An actual experiment with a Velodyne HDL-32 LIDAR is performed, in which a series of scans are taken from at least three basketballs as simple features in the LIDAR proximity. The pre-processing steps and the general estimation evaluation for these experiments are analyzed in section \ref{ch7:experimental}. Two types of sensors are employed in these experiments; first the Velodyne VLP-32 LIDAR is used to verify the TLS solution for the point-cloud registration problem, and second a multi-camera Optitrack system is utilized, which has much better accuracy than the LIDAR in terms of tracking the 3-dimensional positions. The vision system's accuracy is sub-millimeter for each $x$, $y$ and $z$ dimension, which is orders of magnitude better than the estimated position from the LIDAR derived estimates. The measurement covariance for the LIDAR is obtained using a static configuration; the mean and covariance are computed using a simple sampling approach. The Optitrack system is utilized as ground truth for the pose estimates from the LIDAR data.

The minimum number of three features is selected as the centroids of the three spheres in the LIDAR field of view. Also, the features should be visible in at least three cameras from the Optitrack system to be able to accurately capture the ground truth data. This ground truth data is required for the LIDAR pose and the observation vectors provided by the Optitrack system.
For this experimental setup, the LIDAR is static and takes point cloud measurements from surrounding basketballs, which are also static with respect to an inertial coordinate frame on the floor.

In the following sections, the pre-processing steps are discussed in detail.

The experimental results for the point-cloud registration problem in Eq.~\eqref{eqn:cost_ahat_phat} are discussed here.

\begin{figure}[!ht]
    \begin{centering}
      \subfigure[{\bf Optitrack Cameras with Three Spheres as Features and the VLP-32 LIDAR}]
      {\includegraphics[width=0.47\textwidth]{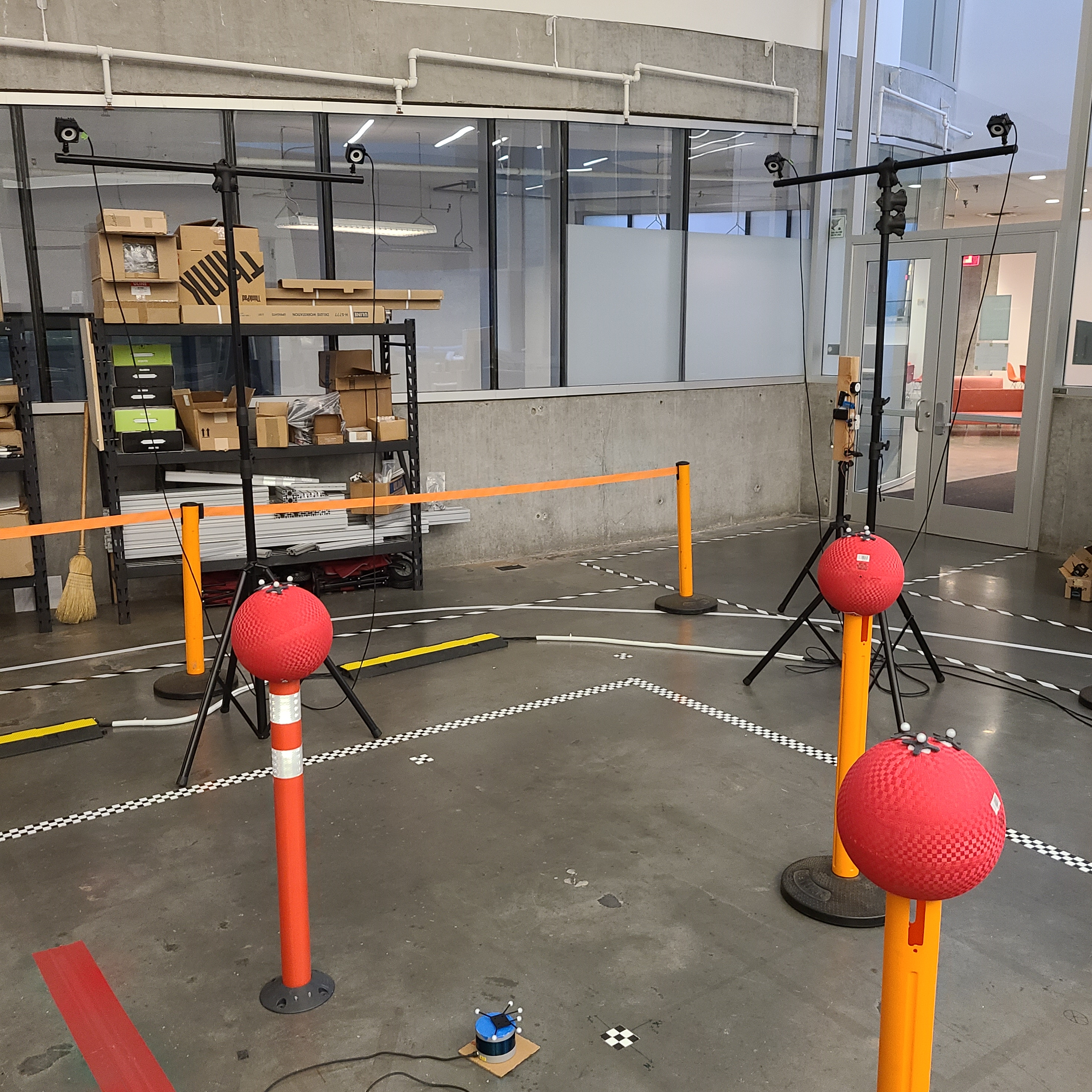}\label{fig:all_in_one_lidar_exp}}
      \subfigure[{\bf Sample rviz Visualization of the Experiments in ROS (Robotics Operating System)}]
      {\includegraphics[width=0.48\textwidth]{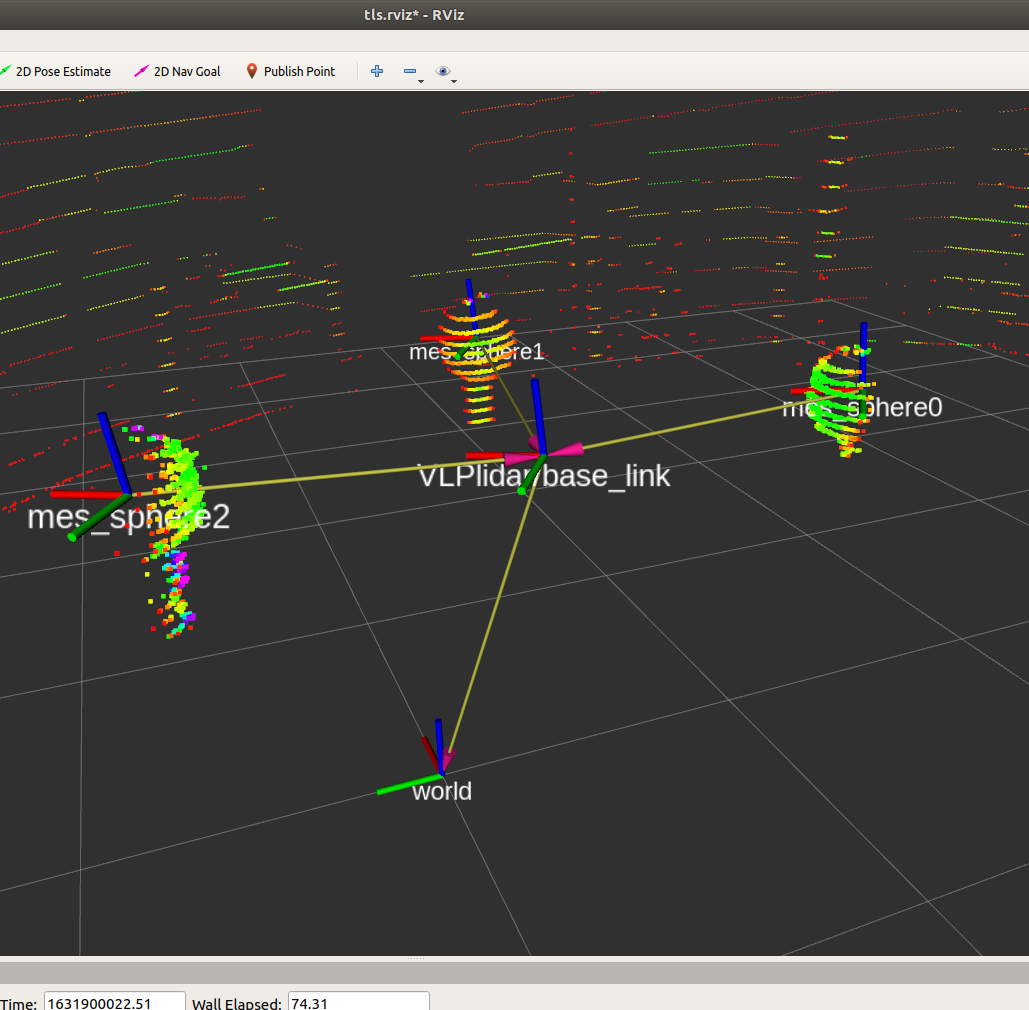}
    \label{fig:ch7_rviz_experiment}}
    \caption{\bf Experimental setup and visualization. }
    \end{centering}
\end{figure}

A picture of the experimental environment is shown in Fig.~\ref{fig:all_in_one_lidar_exp}. The cylinders are there to avoid the ground-floor outliers in LIDAR raw data.
The rviz visualization of the experiments is shown in Fig.~\ref{fig:ch7_rviz_experiment}. 

The Optitrack system follows the rays from a series of markers attached to the objects of interest in the environment. The red-green-blue rods show the coordinate frames on each sphere as well as the LIDAR. The scattered points in the background are the outlier and noisy data from other unwanted objects. In the case of tracking the orientation of an object, the Optitrack software needs at least three markers attached to the object. For each basketball and the LIDAR, four markers are attached to satisfy the tracking requirements of the Optitrack software.

\begin{figure}[!ht]
    \includegraphics[width=\textwidth]{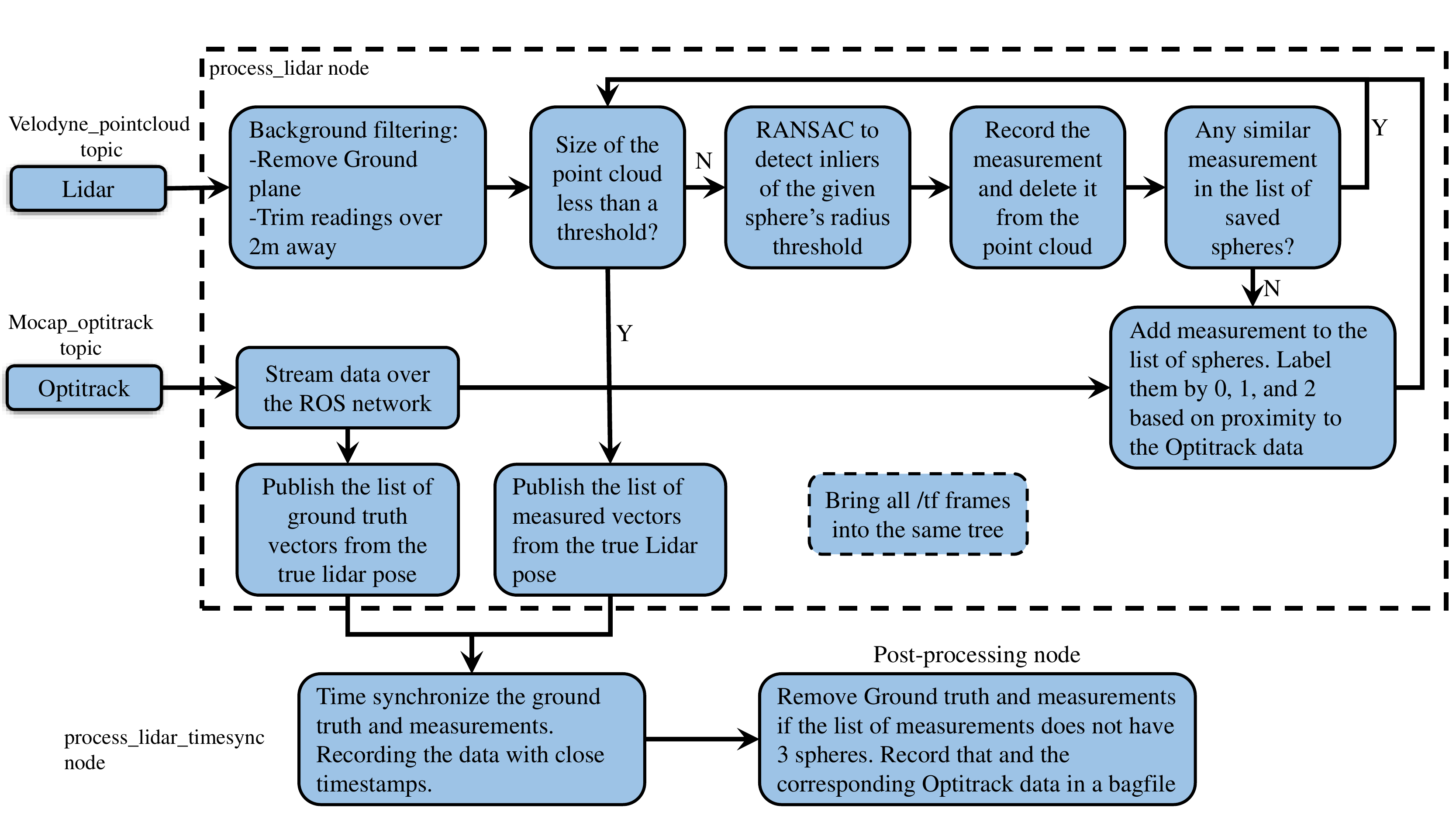}
    \caption{Block diagram of the pre-processing steps for LIDAR and optitrack data before feeding the measurements into the core pose estimation algorithm.}
    \label{fig:ch7_preprocess_diag}
\end{figure}

In this experimental setup, the LIDAR and the basketballs are not moving with respect to a pre-determined inertial frame on the floor.
It is noted that the ground plane appears in the LIDAR scans and needs to be removed as a pre-processing step to reject the unwanted data for our pose estimation purpose.
The pre-processing steps are elaborated in Fig.~\ref{fig:ch7_preprocess_diag}.

\begin{figure}[!ht]
  \begin{centering}
    \subfigure[{\bf Attitude Errors}]
    {\includegraphics[width=0.48\textwidth]{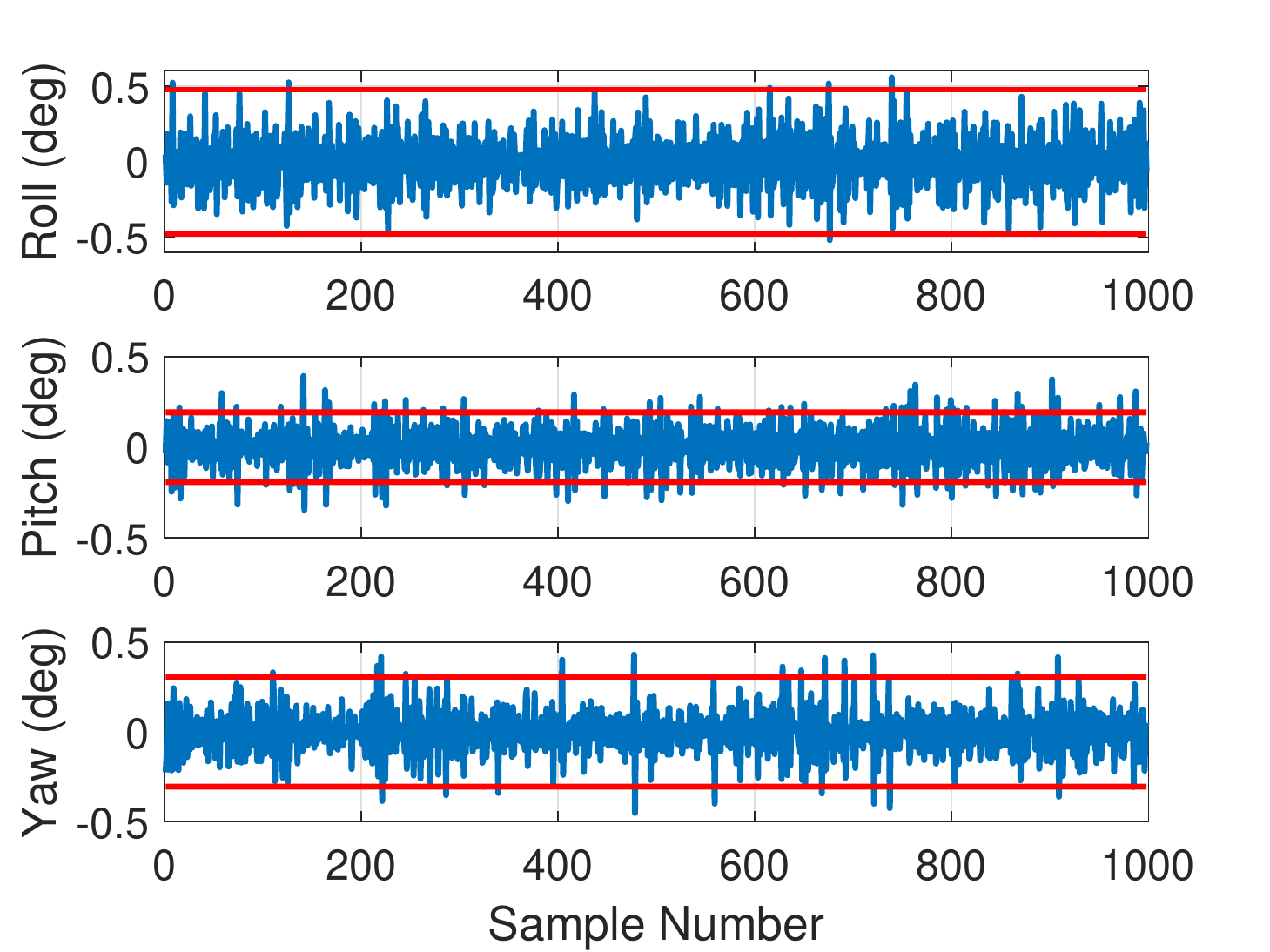}\label{fig:rpy_planar_exp}}
    \subfigure[{\bf Position Errors}]
    {\includegraphics[width=0.49\textwidth]{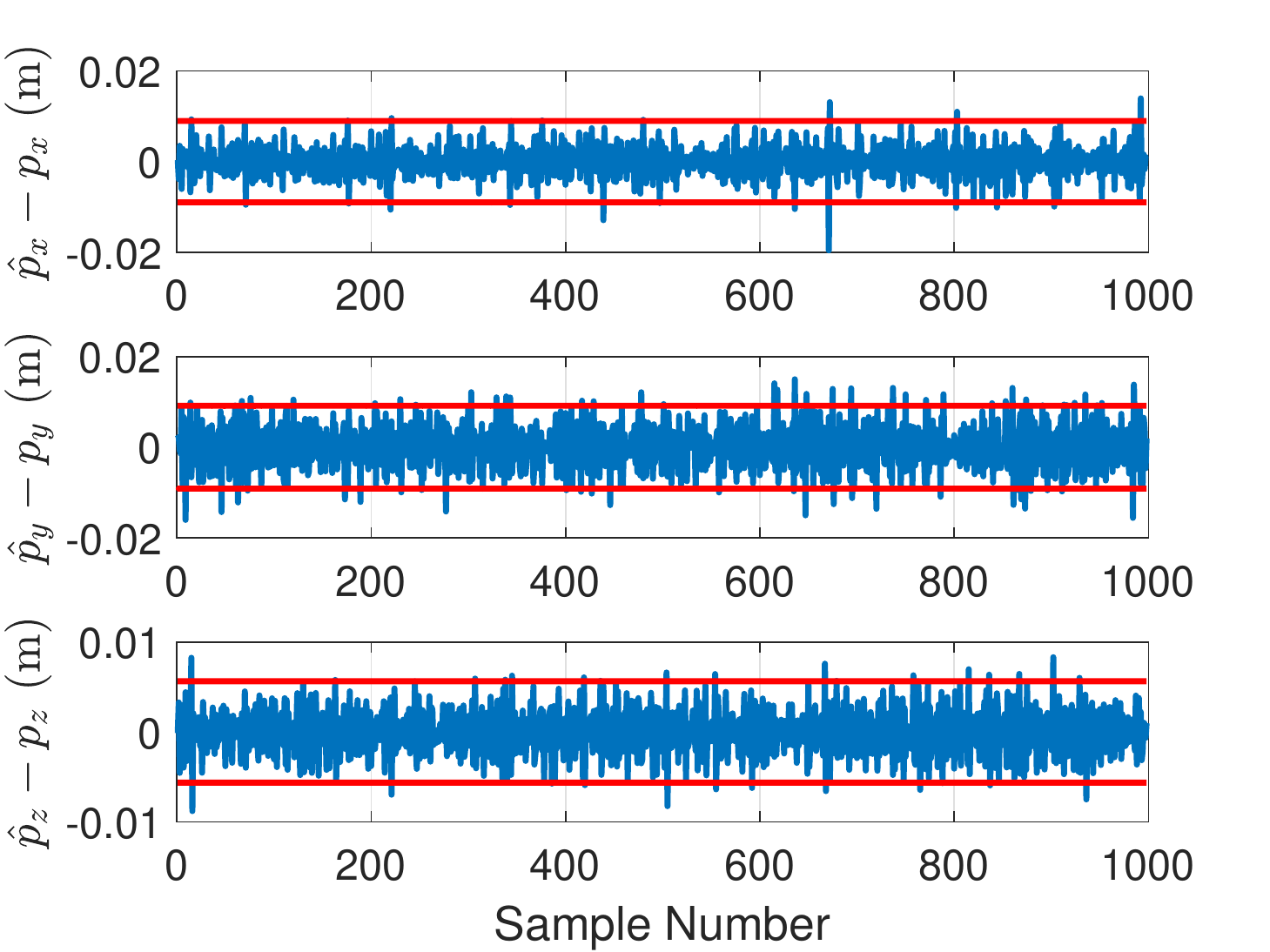}\label{fig:p_planar_exp}}
    \caption{Static LIDAR experiment for the attitude matrix and translation vector.}
  \end{centering}
\end{figure}

\begin{figure}[!ht]
  \begin{centering}
    \subfigure[{\bf Estimation Errors}]
    {\includegraphics[width=0.49\textwidth]{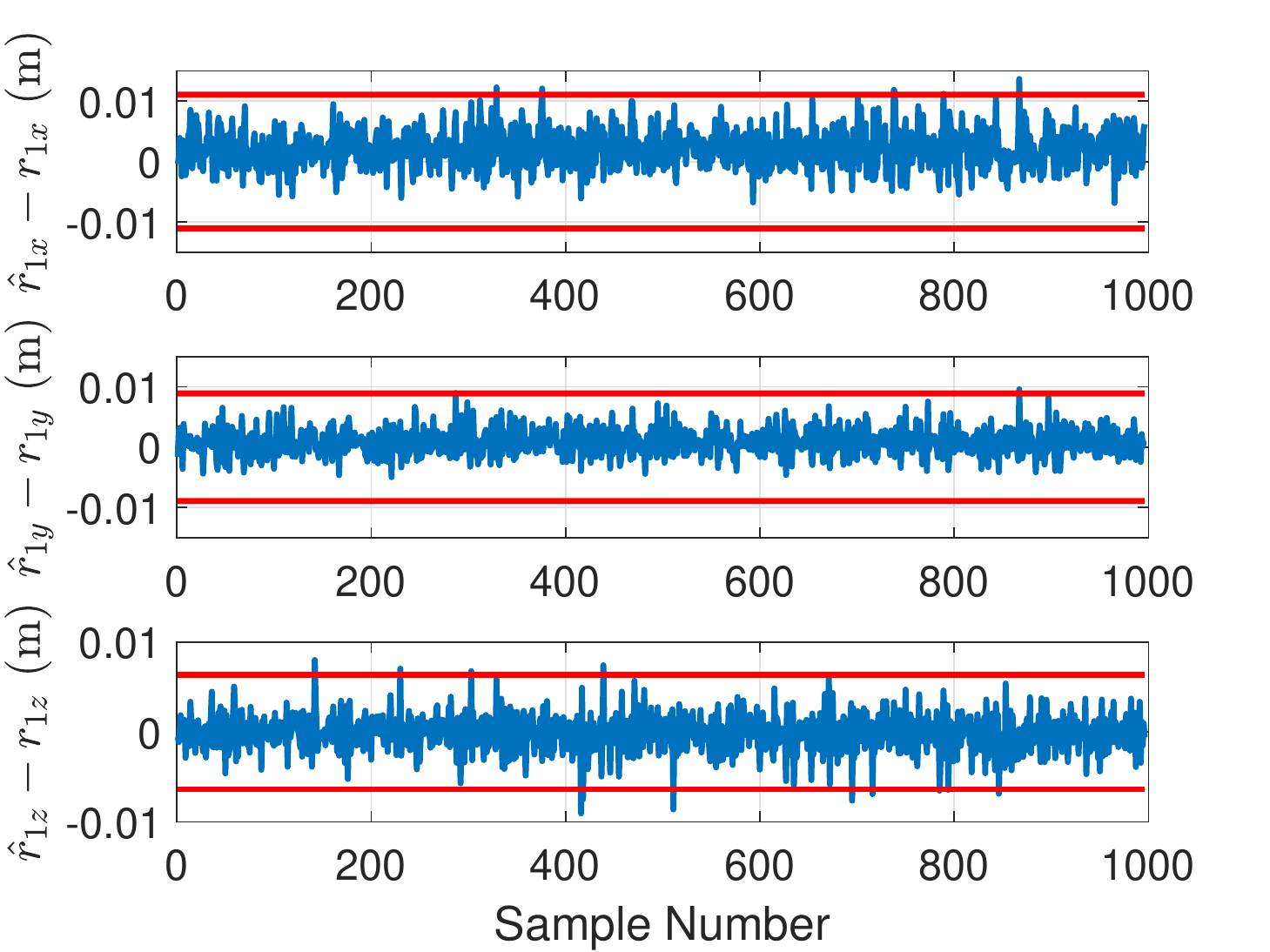}\label{fig:r_est_planar_exp}}
    \subfigure[{\bf  Residual Errors}]
    {\includegraphics[width=0.49\textwidth]{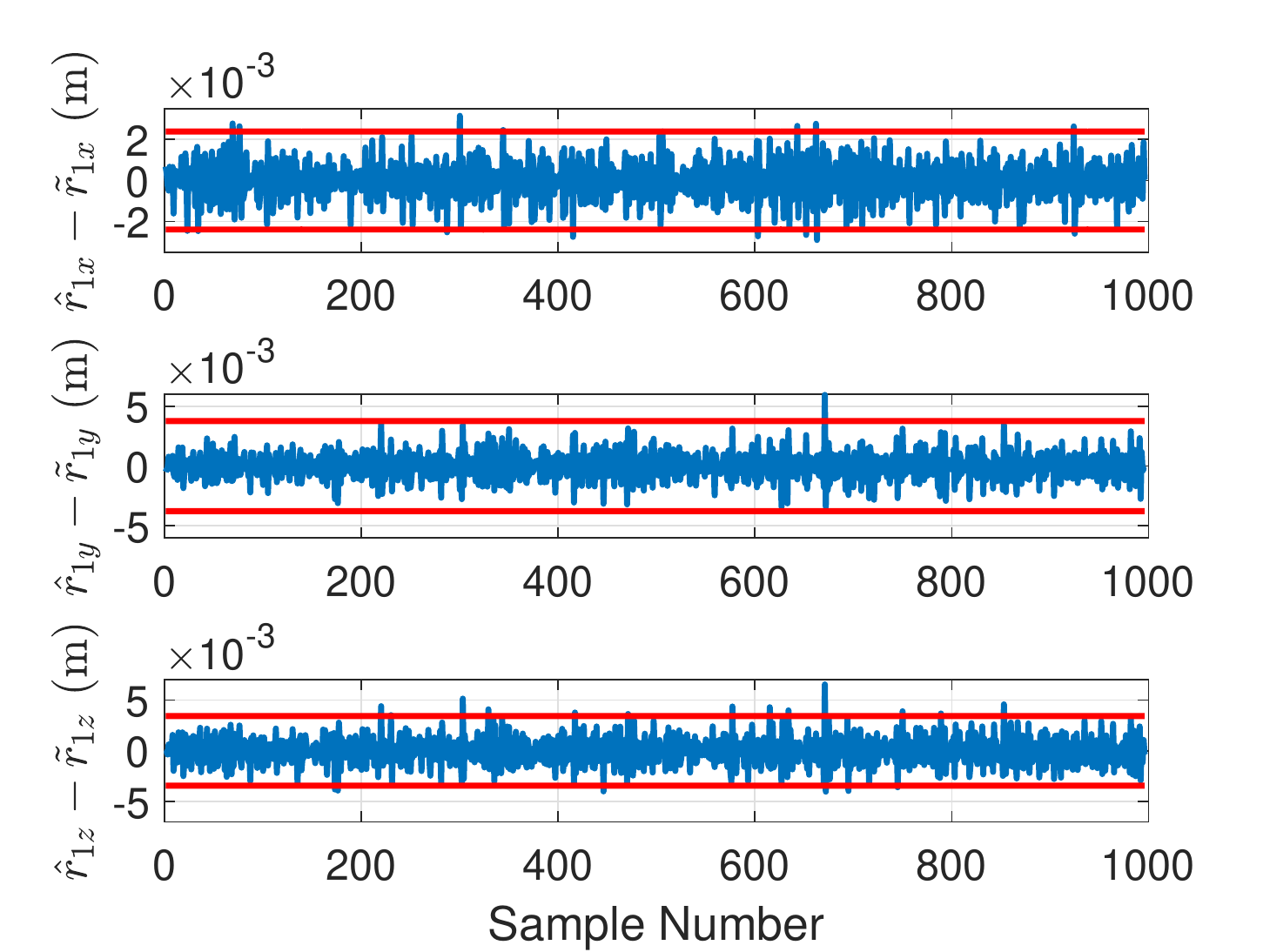}\label{fig:r_res_planar_exp}}
    \subfigure[{\bf Estimation Errors}]
    {\includegraphics[width=0.49\textwidth]{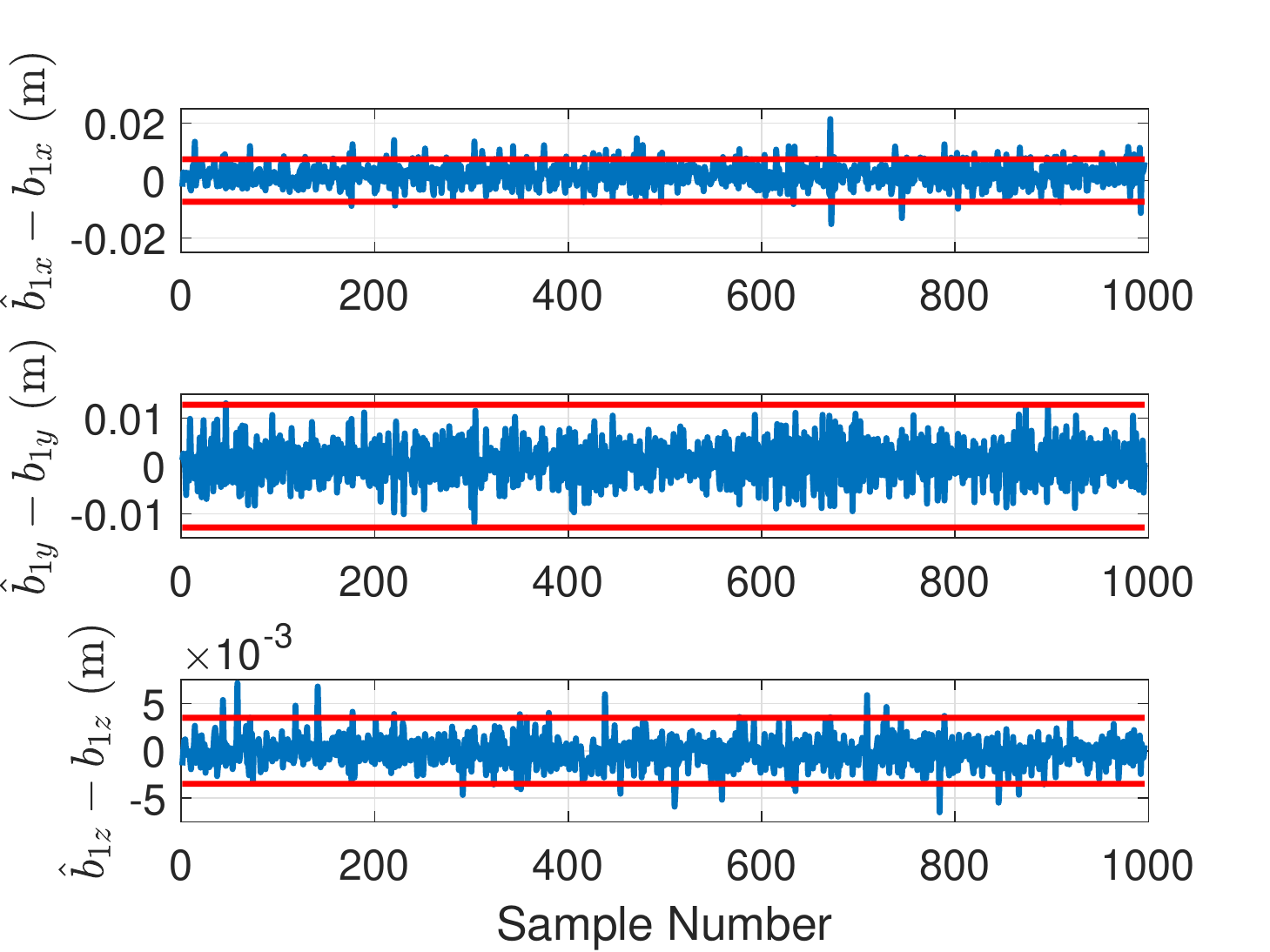}\label{fig:b_est_planar_exp}}
    \subfigure[{\bf Residual Errors}]
    {\includegraphics[width=0.49\textwidth]{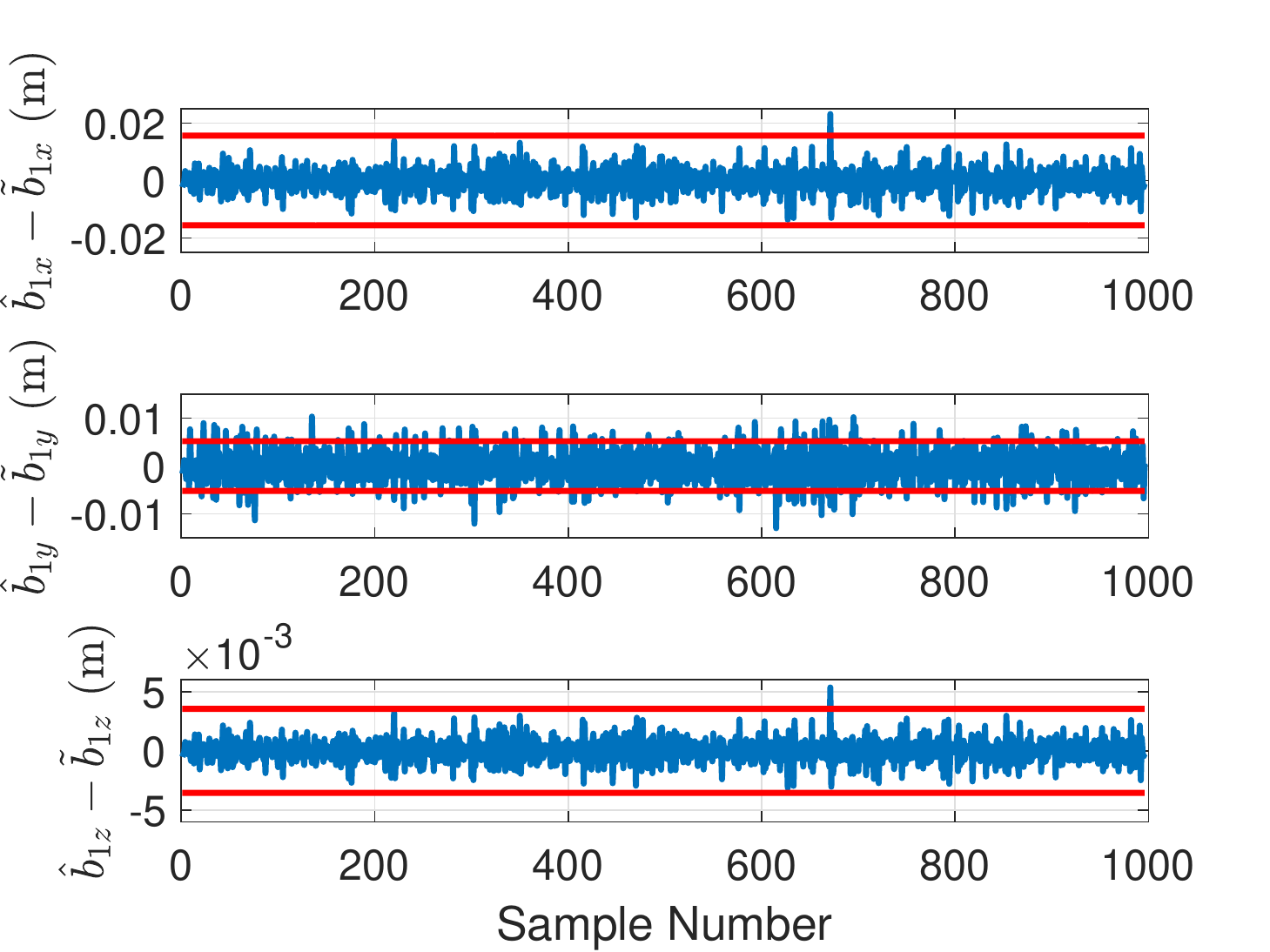}\label{fig:b_res_planar_exp}}
    \caption{Static LIDAR experiment for the observation vectors $\bm{r}_1$ and $\bm{b}_1$, including errors and residuals.}
  \end{centering}
\end{figure}

The results of the TLS pose estimation for this scenario are shown in Figs.~\ref{fig:rpy_planar_exp}, \ref{fig:p_planar_exp}, as well as Figs.~\ref{fig:r_est_planar_exp} to \ref{fig:b_res_planar_exp}.
As seen in the attitude-error plots in Fig.~\ref{fig:rpy_planar_exp}, the angle estimation errors are bounded within limits defined by the 3$\sigma$ bounds, extracted from the attitude error-covariance in Eq.~\eqref{eqn:P_delta_alpha_final}. It is expected that approximately 3 out of 1,000 of the errors should be outside of the 3$\sigma$ bounds. It is seen that there are more than 3 outside of their bounds, which may be due to calibration errors. Still, reasonable performance is given. 

The position errors in $x$, $y$ and $z$ are within their corresponding 3$\sigma$ bounds, originating from the position estimate error-covariance in Eq.~\eqref{calc_cov_phat}.

Figures \ref{fig:r_est_planar_exp} to \ref{fig:b_res_planar_exp} showcase the errors for the observation-vector estimates of $\bm{r}_1$ and $\bm{b}_1$ for the first sphere. Some of the estimates are biased, which again may be do to calibration errors. It is also know that the LIDAR measurement error is non-Gaussian in nature, which may also explain the biased errors. 
The estimate errors for the observation vectors are reasonably bounded by their 3$\sigma$ bounds from Eqs.~\eqref{eqn:cov_ri} and \eqref{eqn:cov_bi} for the residuals, and Eqs.~\eqref{eqn:P_bi} and \eqref{eqn:P_ri} for the estimate errors.
From Section \ref{tls_derivation} the TLS solution has an extra capability of providing estimates for the vector observations, as seen in Figures \ref{fig:r_est_planar_exp} and \ref{fig:b_est_planar_exp}, which are the positions of the vehicle with respect to the landmark features in the environment, as well as the unknown pose.
This ability makes the TLS a solution for the SLAM problem that does not exist in other pose estimation solutions in the literature.
In particular, measurement residuals that fall significantly outside a prescribed error-bound can be used to assess whether the measurements should be used or removed in the estimation process.

\section{Conclusions}
This paper develops an analytical solution for an efficient pose estimation problem within the first-order approximation of estimate errors. Pose estimation is central to simultaneous localization and mapping (SLAM) problems. Efficient estimation makes the controller policy easier to track the desired signals to obtain a more accurate estimate of the states.
The static SLAM problem is shown to be solved as a total least squares (TLS) problem.
A quadratic cost function based on the TLS formulation is introduced for taking into account the attitude matrix and translation vector. The weight matrix in the cost function is extracted from the most generic positive-definite fully populated matrix to include the correlations between the observation vectors in the most general case.
The cost function is then written in an attitude-only format. The error-covariance expression of the attitude error is provided within the small-angle assumption. This assumption leads to a second-order approximation of the cost function in terms of the attitude error. Covariance expressions for the translation vector as well as the estimates and residuals of the observation vectors are obtained analytically. The Fisher information matrix (FIM) is derived, and the error-covariance expressions are proven to be the block inverses of FIM, proving the equality in the Cram\'er-Rao lower bound and thus the efficiency of estimates within first-order attitude errors. Also, an advantage of the proposed solution is that it provides an estimate of the residual for the observation vectors as well as their analytically-derived covariances. These quantities are useful to assess the performance of the pose estimation solution when no access to the ground truth data in an actual SLAM application is available.
Two verification steps are employed to assert the efficacy of the pose estimation approach.
First, a simulation showcases the efficacy of the covariance analyses by simulating observation vectors in a pose estimation problem with 10,000 Monte Carlo samples. 
Second, an actual LIDAR takes scans of at least three surrounding features, and the pipeline generates the pose estimates of the LIDAR as well as residual estimates for the observations.
The Optitrack multi-camera system is utilized for scanning the ground truth data of the sensor's position and orientation. The estimate error-covariances are verified to their corresponding estimate or residual errors in the actual experimental setup, which validates the derived analytical expressions.

\section*{Appendix}
\renewcommand{\theequation}{A\arabic{equation}}\setcounter{equation}{0}
A list of some preliminary equations used in the paper is shown here.
For the cross product of two $3\times 1$ vectors $\bm{a}$ and $\bm{b}$, the following relation is given:
\begin{align}
    \bm{a}\times\bm{b}=[\bm{a}\times]\bm{b}=-\bm{b}\times\bm{a}=-[\bm{b}\times]\bm{a}\label{eqn:cross_prod_def}
\end{align}
The cross product matrix of a vector $\bm{a}$ is a skew-symmetric matrix, so that
\begin{align}\label{eqn:skew-symmetric_cross_prod}
    [\bm{a}\times]^T=-[\bm{a}\times]
\end{align}
where $[\bm{a}\times]$ denotes the $3\times3$ cross product matrix constructed from the vector $\bm{a}$ as
\begin{align}\label{cross_prod_def}
    [\bm{a}\times]=\begin{bmatrix}0 & -a_3 & a_2\\
    a_3 & 0 & -a_1\\
    -a_2 & a_1 & 0\end{bmatrix}
\end{align}
The Kronecker product \cite{steeb2011matrix} also can be combined with the vec operator in the following identity, where the
notation vec($\cdot$) \cite{1084534} for an $m\times n$ matrix $A$ results in an $(mn)\times 1$ matrix vec$(A)$ that is constructed by stacking the columns of $A$:
\begin{equation}\label{eqn:kronecker_vec}
    A \bm{z}=(\bm{z}^T \otimes I_{m\times m})\text{vec}(A)
\end{equation}

\bibliography{sample}

\end{document}